\documentclass[lettersize,journal]{IEEEtran}
\usepackage{amsmath,amsfonts}
\usepackage{algorithmic}
\usepackage{algorithm}
\usepackage{array}
\usepackage[caption=false,font=normalsize,labelfont=sf,textfont=sf]{subfig}
\usepackage{textcomp}
\usepackage{stfloats}
\usepackage{url}
\usepackage{verbatim}
\usepackage{graphicx}

\usepackage{cite}
\usepackage{booktabs}
\usepackage{multirow}
\usepackage[table]{xcolor}
\usepackage[hidelinks]{hyperref}

\begin{document}

\title{SR-Stereo \& DAPE: Stepwise Regression and Pre-trained Edges for Practical Stereo Matching}

\author{Weiqing Xiao, Wei Zhao$^{\ast}$
        % <-this % stops a space
\thanks{\textit{Corresponding author: Wei Zhao.}}
\thanks{Weiqing Xiao and Wei Zhao are with 
the School of Electronic and Information Engineering, 
Beihang University, Beijing 100191, 
China (e-mail: xiaowqtx@buaa.edu.cn; zhaowei203@buaa.edu.cn)}
}

\markboth{Journal of \LaTeX\ Class Files,~Vol.~14, No.~8, August~2021}%
{Shell \MakeLowercase{\textit{et al.}}: A Sample Article Using IEEEtran.cls for IEEE Journals}

\IEEEpubid{0000--0000/00\$00.00~\copyright~2021 IEEE}

\maketitle

\begin{abstract}
  Due to the difficulty in obtaining real samples and ground truth, 
  the generalization performance and domain adaptation performance are 
  critical for the feasibility of stereo matching methods in practical 
  applications. However, there are significant distributional discrepancies 
  among different domains, which pose challenges for generalization and domain 
  adaptation of the model. Inspired by the iteration-based methods, 
  we propose a novel stepwise regression architecture. This architecture 
  regresses the disparity error through multiple 
  range-controlled clips, which effectively 
  overcomes domain discrepancies. We implement this architecture 
  based on the iterative-based methods , and refer to this new 
  stereo method as SR-Stereo. Specifically, a new stepwise regression 
  unit is proposed to replace the original update unit in 
  order to control the range of output. Meanwhile, a regression 
  objective segment is proposed to set the supervision 
  individually for each stepwise regression unit. In addition, 
  to enhance the edge awareness of models adapting new domains 
  with sparse ground truth, we propose Domain Adaptation based on Pre-trained 
  Edges (DAPE). In DAPE, a pre-trained stereo model and an edge estimator 
  are used to estimate the edge maps of the target domain images, which 
  along with the sparse ground truth disparity are used to fine-tune 
  the stereo model. The proposed SR-Stereo and DAPE are extensively 
  evaluated on SceneFlow, KITTI, Middbury 2014 and ETH3D. 
  Compared with the SOTA methods and generalized methods, the proposed 
  SR-Stereo achieves competitive in-domain and cross-domain performances. 
  Meanwhile, the proposed DAPE significantly improves the performance 
  of the fine-tuned model, especially in the texture-less and detailed regions.
  The code is available at \href{https://github.com/zhuxing0/SR-Stereov1-DAPE}{https://github.com/zhuxing0/SR-Stereov1-DAPE}.
\end{abstract}


\section{Introduction}
\IEEEPARstart{D}{epth} information is key to computer vision 
and graphics research in the real world. Accurate depth 
estimation is vital for fields such as autonomous driving, 
robot navigation and action recognition. Common depth estimation 
methods include structured light~\cite{zhang2013real}, 
stereo matching~\cite{liu2024guard, li2021revisiting, chang2018pyramid, guo2019group, xu2022attention}, 
and radar~\cite{feng2018lane}, etc. 
The structured light method projects a coded pattern 
onto the object surface and estimates the depth by observing 
the distortion of the pattern imaged on the object surface. 
Due to the need to project multiple patterns consecutively, 
structured light methods are generally only suitable for 
static indoor scenes. The radar obtains a sparse depth map 
by transmitting pulses and receiving echoes. Millimeter-wave 
radar has a long detection range but fuzzy spatial resolution, 
while laser radar has a high spatial resolution but is expensive. 
In contrast, stereo matching calculates the depth map by estimating 
the horizontal displacement map (i.e., the disparity) of the 
pixels between the corrected left and right image pairs. 
By utilizing only two cameras, stereo matching is able to provide 
dense depth maps in different scenes, and is therefore considered 
a cost-effective and widely applicable method for depth estimation.

\begin{figure}[t]
  \centering
  \includegraphics[width=0.485\textwidth]{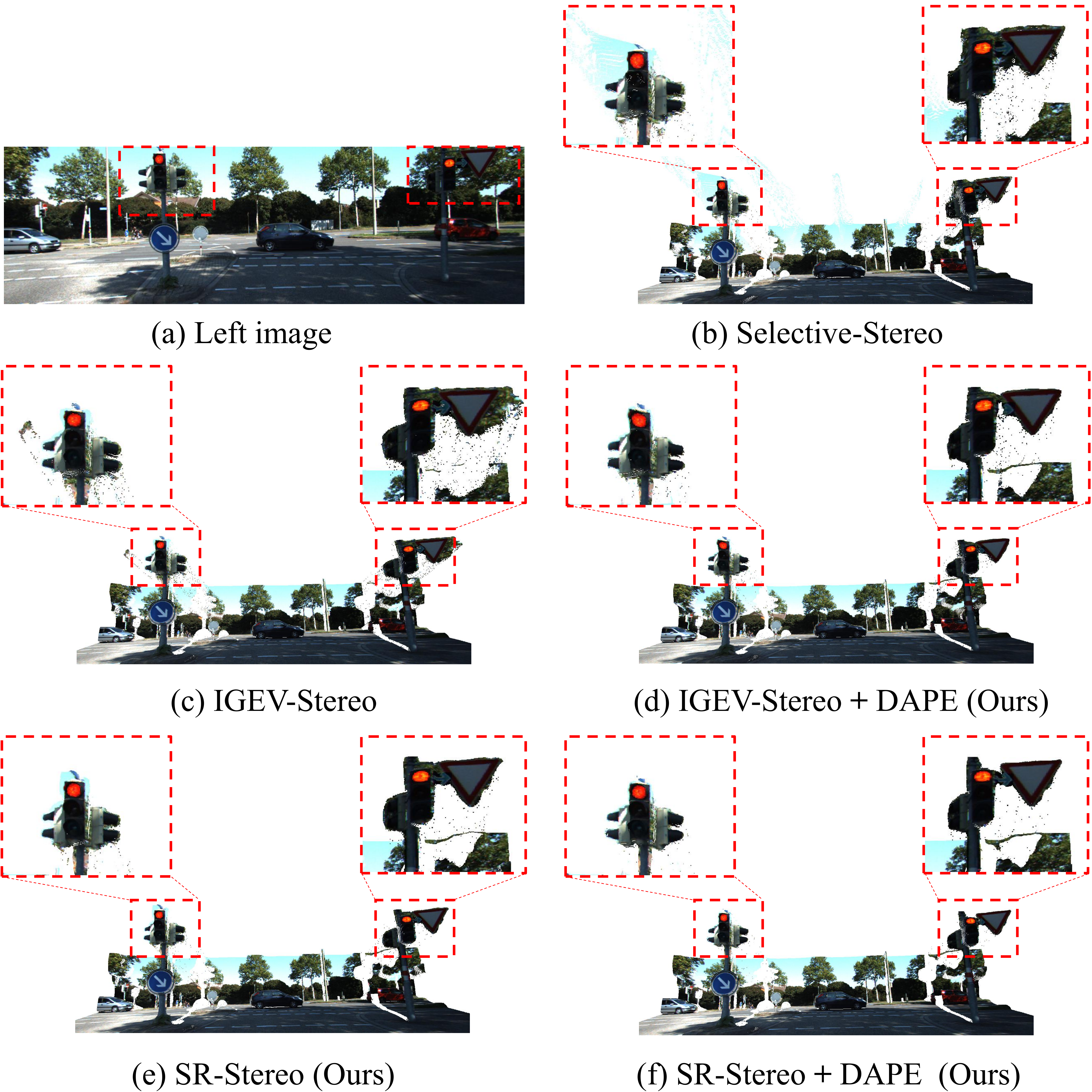}
  \caption{Comparison of reconstructed point clouds on KITTI. 
  All methods are trained on SceneFlow and fine-tuned on KITTI. 
  During inference, all methods run 15 disparity updates. 
  Our SR-Stereo performs better in the detailed regions. 
  In addition, the proposed fine-tuning framework 
  DAPE effectively improves the performance of 
  existing methods fine-tuned on sparse ground truth.}
  \label{fig:1}
\end{figure}

\IEEEpubidadjcol

Many learning-based stereo methods~\cite{chang2018pyramid,
li2021revisiting,guo2019group,xu2022attention,lipson2021raft,li2022practical,
xu2023iterative,zhao2023high,wang2024selective,weinzaepfel2023croco} 
have achieved encouraging success. 
Some works~\cite{chang2018pyramid,guo2019group,xu2022attention} 
focus on designing the 4D cost volume, which 
characterizes the cost of pixel matching between 
left and right image pairs. They use 3D convolutional networks 
to aggregate and regularize the entire cost volume, and regress 
it to obtain the disparity. Recently, 
iteration-based methods~\cite{lipson2021raft,li2022practical,
xu2023iterative,zhao2023high,wang2024selective,
weinzaepfel2023croco} have shown great 
potential in terms of both accuracy and real-time performance, 
and thus have become the mainstay of current research. 
The iteration-based methods use similarity features 
from the cost volume to predict the disparity error 
for optimization of the disparity. 
This approach avoids the expensive cost of 
aggregating and regularizing cost volume and achieves a 
balance between performance and efficiency by controlling 
the number of iterations. 

However, real training samples and ground truth is usually 
difficult to obtain for stereo methods in practical applications. 
Existing methods~\cite{shen2023digging,zhang2019ga,zhang2020adaptive,
zeng2021deep} are generally pre-trained 
on a large synthetic dataset (e.g., SceneFlow~\cite{mayer2016large}) 
and then fine-tuned or directly inferred on real datasets, 
which makes higher demands on the generalization performance 
and domain adaptation performance.
In addition, there are significant distributional discrepancies 
among different datasets due to differences in scene and resolution. 
For example, the maximum disparity in the Middlebury~\cite{scharstein2014high} 
exceeds 250 pixels, while all the disparities in the 
KITTI~\cite{geiger2012we,menze2015object} are within 100 pixels. 
These discrepancies make it difficult for a method that performs 
well on one dataset to directly achieve the same performance 
on other datasets. Therefore, a new architecture needs to be 
designed to overcome the domain discrepancies to 
achieve better robustness. 
% Therefore, we need to improve the existing method 
% so that it overcomes the discrepancies across domains.
% has good performance in estimating disparity across domains.

Our architecture is inspired by iteration-based methods. Iteration-based methods 
aim to correct disparity by directly predicting the current disparity error 
using update units. However, influenced by the disparity distribution, 
the disparity errors tend to vary across domains, 
which limits the generalization performance of iteration-based methods. 
Additionally, tens of updates are usually required to obtain accurate results. 
% Since this is the case, why not evenly distribute 
% the predicted error among multiple update units? Therefore
Based on these observations, we propose 
a novel stepwise regression architecture. In this architecture, we split 
the disparity error into multiple segments with fixed ranges and use 
multiple update units to predict them separately. This architecture has 
the following advantages: 1) the segments are range-controlled and 
domain-independent, and thus easy to generalize. 
2) it still requires only tens of updates, and does not introduce additional 
computational and time costs.

We implement this architecture on the basis of iteration-based methods, 
and refer to this new stereo method as SR-Stereo. Specifically, we propose 
a stepwise regression unit to replace the original update unit. The output 
of the stepwise regression unit is range-controlled rather than 
unconstrained as in the original update unit. Then, we propose a 
regression objective segment, which sets a separate regression 
objective for each stepwise regression unit based on the splitting 
result of the disparity error. Further, we introduce 
a Disparity Clip-Balanced Weight to improve the accuracy of the predicted segments in 
the stepwise regression units.

Furthermore, we observe that models fine-tuned with sparse 
ground truth suffer from severe edge blurring, i.e., edge 
disparity errors. As shown in Figure~\ref{fig:1}, even the state-of-the-art 
models~\cite{xu2023iterative, wang2024selective} 
in the KITTI online leaderboard fail to accurately 
estimate the edge disparity. This further limits the 
effectiveness of stereo methods in real-world applications, 
as the vast majority of efficient labeling methods (e.g., radar) 
can only obtain sparse ground truth. The most direct way 
to mitigate this problem is to introduce edge-dependent 
supervision, but current edge feature extraction 
methods~\cite{dong2023egfnet, cao2022pavement, guo2021barnet}
have difficulty in obtaining accurate and concise edge 
information due to the overly sparse disparity ground 
truth and the interference of object textures in RGB images.

In this paper, we propose a novel model fine-tuning framework 
to address the problem of blurring in edge details, which we name 
Domain Adaptation based on Pre-trained Edge (DAPE).
The two key observations behind this framework are: 1) in contrast 
to RGB images, in disparity, the texture of the object is weakened while 
concise edge information is highlighted. 2) for 
the model pre-trained with dense ground truth, 
its predicted disparity on new domains tends to have accurate edge contours. 
Therefore, we first propose a lightweight edge 
estimator that takes both disparity and RGB image as inputs to 
generate a dense edge map. This estimator, 
along with SR-Stereo, is pre-trained on a 
virtual dataset. We utilize the pre-trained 
SR-Stereo to generate the disparity map for the target domain, 
which is further fed into the edge estimator to predict the edge map. 
Then, the background pixels in the edge map are retained, 
while the foreground pixels are filtered out.
Following UCFNet~\cite{shen2023digging}, these pixels are defined as pseudo-labels.
Finally, we use the edge pseudo-labels as additional 
supervision during the model fine-tuning process, which 
effectively improves the model's disparity estimation performance on edge details.

In summary, our main contributions are:

\begin{itemize}
  \item{We propose a novel stereo method called SR-Stereo. It splits the 
  disparity error into multiple segments for regression, which 
  effectively overcomes the domain discrepancies 
  (i.e., differences in disparity distribution). % among different domains
  More importantly, we propose a stepwise regression architecture, 
  which provides a new implementation idea for generalized stereo matching.}

  \item{We propose a novel fine-tuning framework for real datasets, 
  named DAPE. By employing generated edge pseudo-label to 
  supervise the model fine-tuning process, the proposed DAPE 
  effectively improves the disparity estimation performance of 
  models fine-tuned with sparse ground truth.}

  \item{We perform extensive experiments on SceneFlow, KITTI, 
  Middlebury and ETH3D. Compared to the state-of-the-art methods, 
  our SR-Stereo achieves competitive performance on several benchmarks. 
  Meanwhile, compared to the majority of generalized stereo methods, 
  our SR-Stereo achieves the best performance on Middlebury and ETH3D. 
  The generalization performance of our method on KITTI is also very 
  competitive among the methods of the last two years. In addition, 
  experimental results on multiple realistic datasets show that DAPE 
  significantly improves the performance of the fine-tuned model, 
  especially in texture-less and detailed regions.}

\end{itemize}

\section{Related Work}

\subsection{Iterative-based Stereo Matching}

Compared to cost aggregation-based methods, 
iteration-based methods~\cite{zhang2019ga,zhang2020adaptive} 
achieve significant improvement in both efficiency and accuracy. 
RAFT-Stereo~\cite{lipson2021raft} is the first 
iteration-based method that innovatively introduces 
multi-level GRU~\cite{cho2014learning} (Gated Recurrent Unit) 
to update disparities in stereo matching. Based on RAFT-Stereo, 
DLNR~\cite{zhao2023high} replaces GRU with decoupled 
LSTM~\cite{graves2012long} (Long Short-Term Memory) to retain 
more high-frequency information during the iterative process, 
and designs a normalized refinement module to capture more detailed 
information at full resolution. Furthermore, CREStereo~\cite{li2022practical} 
applies the iterative process to different resolutions of disparity 
and proposes a hierarchical refinement network to update disparities 
in a coarse-to-fine manner. IGEV-Stereo~\cite{xu2023iterative} 
constructs a combined geometric encoding volume by combining 
Geometric Encoding Volume and All-Pairs Correlation, and uses a lightweight 3D 
regularization network to regress a rough initial disparity, which 
effectively improves the iterative efficiency and accuracy. 
Selective-Stereo~\cite{wang2024selective} further enhances performance 
by extracting disparity information of different frequencies using 
GRUs with varying kernel sizes. Overall, these methods 
aim to directly regress the entire disparity error to 
achieve fast disparity optimization. In contrast, our SR-Stereo 
does not insist on directly regressing the disparity error, 
but instead splits it into multiple range-controlled and 
domain-independent segments to overcome the distributional discrepancies among different domains.

% Overall, these methods aim to directly 
% regress the disparity error to correct the current disparity. 
% In contrast, our SR-Stereo focuses on overcoming the discrepancies 
% across domains to achieve better generalization and domain adaptation 
% performance. To this end, in SR-Stereo, the disparity error is 
% decomposed into multiple range-controlled segments 
% that are domain-independent for regression.

\subsection{Generalized Stereo Matching and Domain Adaptation}

Several works have focused on 
cross-domain generalization and domain adaptation. 
UCFNet~\cite{shen2023digging} narrows domain discrepancies 
by refining disparities in stages and adaptively adjusts the 
disparity search space at each stage through uncertainty estimation. 
Additionally, it proposes a fine-tuning framework based on 
pseudo-labels of target domain disparities to enhance the 
model's domain adaptation performance. HVT~\cite{chang2023domain} 
enriches the distribution of training data by transforming a 
synthetic dataset at three levels (global, local, and pixel), 
thereby improving the model's generalization ability. 
GraftNet~\cite{liu2022graftnet} incorporates broad-spectrum 
features from a large-scale dataset into stereo matching to 
enhance the model's robustness to image styles. 
Rao et al.~\cite{rao2023masked} enhance the learning of 
structural information by integrating stereo matching and 
image reconstruction tasks, thereby improving the model's 
generalization performance. Zhang et al.~\cite{zhang2022revisiting} 
explicitly constrain the feature consistency of matching pixel 
pairs by introducing a stereo contrastive feature loss function. 
DKT~\cite{zhang2024robust} introduces a teacher-student architecture, 
which analyzes the differences between ground truth labels and 
pseudo-labels to enhance the model's robustness. In contrast, 
our SR-Stereo focuses on constructing domain-independent regression objectives, 
specifically by predicting range-controlled disparity error segments 
to achieve more robust performance.

In addition, making good use of the few real 
samples is important for domain adaptation of the model. 
Some works~\cite{xu2023iterative,lipson2021raft,wang2024selective} 
have demonstrated that even using only a few samples, 
the performance of pre-trained models in a new domain 
can be significantly improved. 
However, these methods simply fine-tune 
the model using the ground truth 
disparity of the new domain, without considering 
the impact of ground truth density on domain adaptation.  
For instance, models fine-tuned with sparse ground truth 
often suffer from severe edge blurring. 
EdgeStereo~\cite{song2020edgestereo} mitigates this issue by integrating stereo matching and 
edge detection tasks. Our DAPE provides a simpler solution by using a 
pre-trained stereo model and edge estimator to offer additional edge supervision for 
model fine-tuning. 
% by offering 
% additional edge supervision through a pre-trained stereo model 
% and edge estimator. 
In contrast to EdgeStereo, the stereo model 
and edge estimator in DAPE are completely decoupled, thus 
avoiding unnecessary overhead during the inference.

\section{METHODOLOGY}

\subsection{Overview}\label{Overview}

In this chapter, we provide a novel and complete solution for 
cross-domain generalization and domain adaptation for stereo methods. 
Firstly, we propose a novel stereo method, 
SR-Stereo, which overcomes the domain discrepancies 
by regressing multiple range-controlled 
disparities. Then, we propose a general and reliable 
framework, DAPE, for domain adaptation on sparse ground truth. 

Objectively, SR-Stereo and DAPE are independent, i.e., they 
can be used independently to solve specific problems. 
In Section~\ref{Preliminaries}, we describe the principle of existing iteration-based methods 
and highlight the essential improvements introduced by the proposed SR-Stereo. 
In Section~\ref{SRSTEREO} and Section~\ref{DAPE}, we describe the proposed SR-Stereo and DAPE in detail.

\begin{figure}[t]
  \centering
  \includegraphics[width=0.49\textwidth]{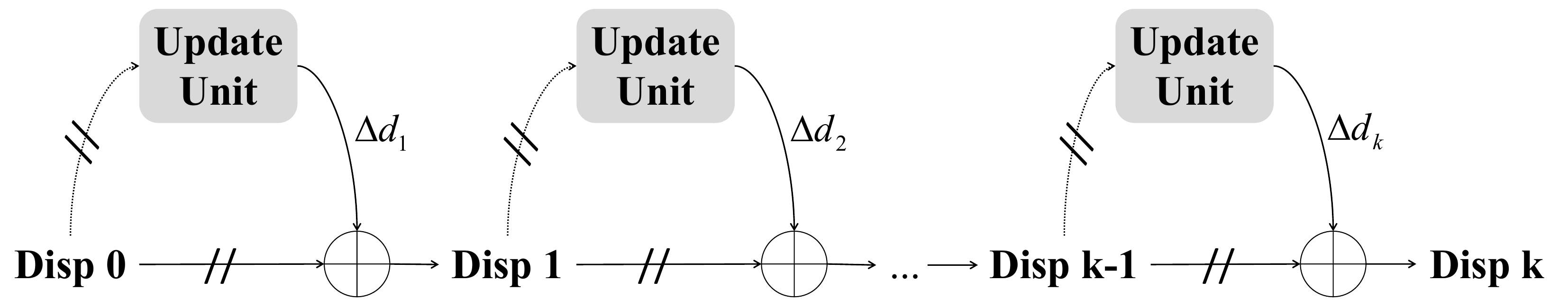}
  \caption{The disparity update process of 
  iteration-based methods. In this process, 
  the update unit outputs the residual disparity 
  to update the current disparity. The $//$ means 
  stop gradient.}
  \label{fig:14}
\end{figure}

\subsection{Preliminaries}\label{Preliminaries}
Distinct from the cost aggregation-based methods, the iteration-based methods 
employ update units to predict and optimize the disparity. 
Ignoring steps such as upsampling, the disparity update can be roughly 
summarized by the process in Figure~\ref{fig:14}. Obviously, the whole process can 
be formulated as:
\begin{equation}
  \label{loss_iteration0}
  d_{k} = d_{k-1}^* + \Delta d_{k}
\end{equation}
where $d_{k}$ denotes the predicted disparity after $k$ iterations, 
$\Delta d_{k}$ denotes the residual disparity in the $k$-th iteration, and * denotes 
no gradient. Then, during training, all the predicted disparities are supervised:
\begin{equation}
  \label{loss_iteration}
  Loss = \sum_{k=1}^{N} \gamma_{k} ||d_{gt}-d_{k}||
\end{equation}
where $Loss$ denotes the training loss, 
$d_{gt}$ denotes the ground truth disparity, 
$\gamma_{k}$ is the loss weight (hyperparameter) 
and $||.||$ denotes common loss functions (e.g., L1, SmoothL1, etc.). 
Considering Eq~\ref{loss_iteration0}, 
the supervision in Eq~\ref{loss_iteration} can be equated:
\begin{equation}
  Loss = \sum_{k=1}^{N} \gamma_{k} || d_{k-1}^{error} - \Delta d_{k} ||
\end{equation}
in which
\begin{equation}
  d_{k-1}^{error} = d_{gt} - d_{k-1}^*
\end{equation}

\begin{figure*}[t]
  \centering
  \includegraphics[width=\textwidth]{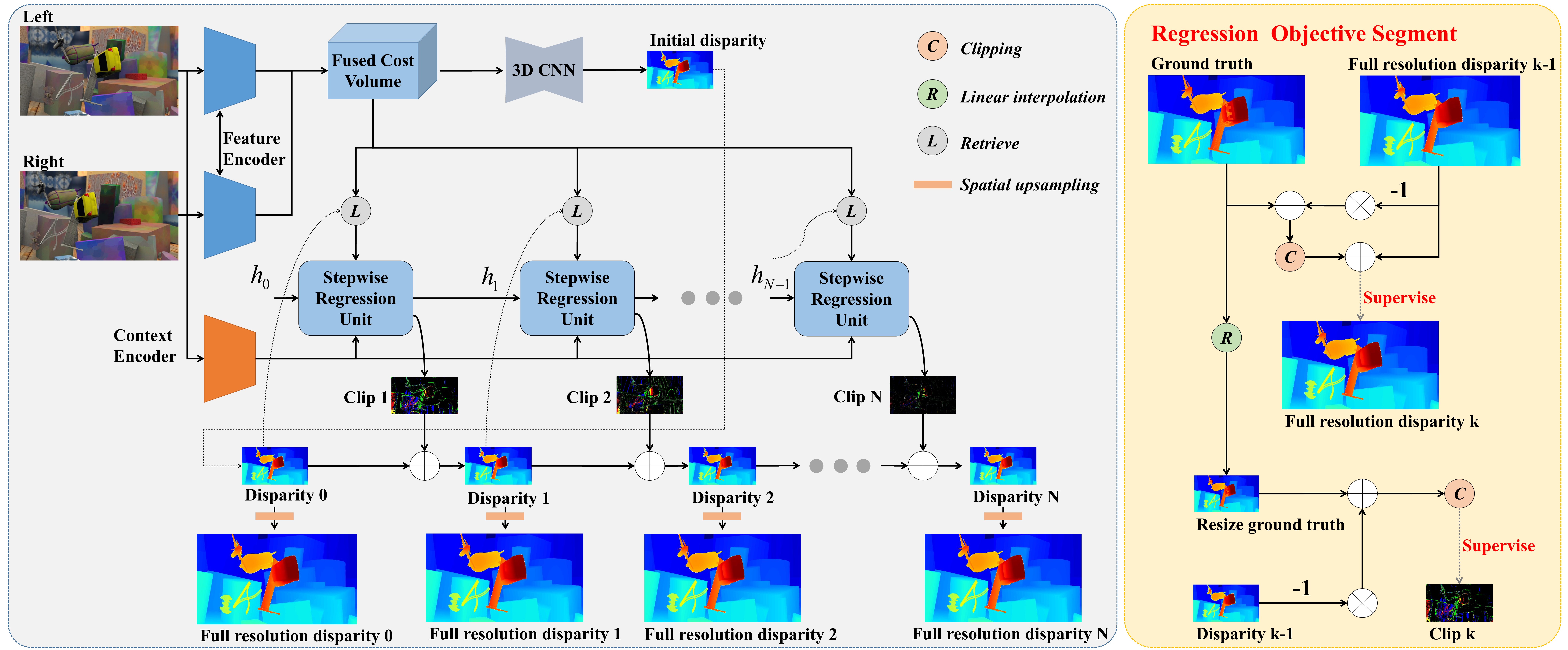}
  \caption{The overall architecture of the proposed SR-Stereo. 
  Compared to iteration-based methods, SR-Stereo is specially designed 
  in terms of the update unit and the regression objective. Specifically, 
  we propose a stepwise regression unit that outputs range-controlled 
  disparity clips, rather than unconstrained residual disparities. 
  Further, we design separate regression objectives for each stepwise 
  regression unit, instead of simply using the disparity error.}
  \label{fig:3}
\end{figure*}

\begin{figure}[t]
  \centering
  \includegraphics[width=0.49\textwidth]{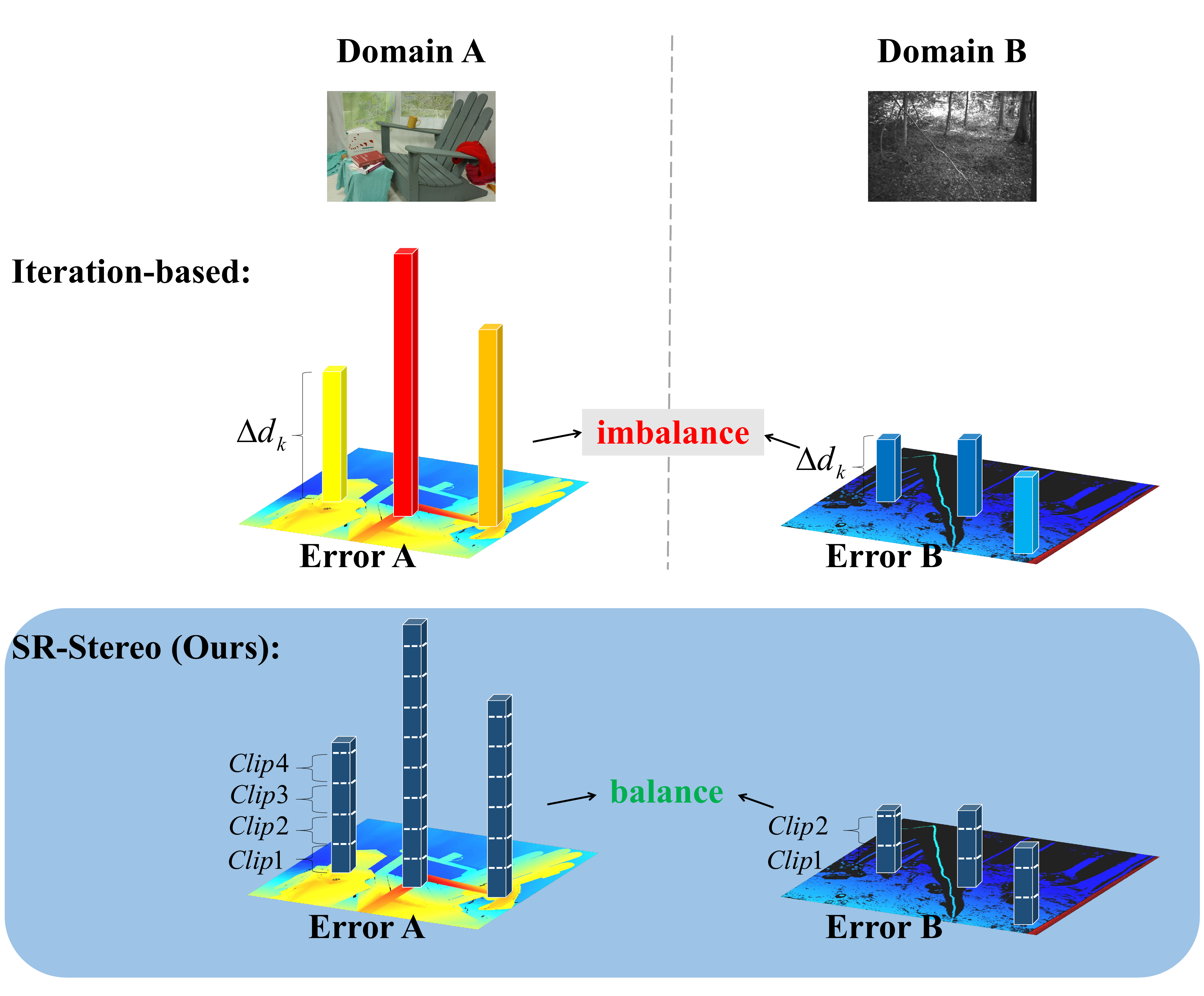}
  \caption{Visualization of the regression objectives of SR-Stereo and 
  iteration-based methods. The iteration-based methods regress disparity 
  error by predicting residual disparity $\Delta d_{k}$, while SR-Stereo splits the disparity 
  error into multiple segments and regresses them by predicting 
  multiple disparity clips.}
  \label{fig:15}
\end{figure}

Essentially, the core of the iteration-based methods is the regression 
of the residual disparity $\Delta d_{k}$ to the disparity error $d_{gt} - d_{k-1}^*$. 
In contrast, in our SR-Stereo, the disparity error is split into 
multiple range-controlled segments for regression, thus 
mitigating the domain discrepancies, as shown in Figure~\ref{fig:15}. 
In this paper, we reconstruct the update unit and regression 
objective in existing iteration-based methods to implement SR-Stereo. 
To highlight the differences, we refer to the output of the 
new update unit as a disparity clip rather than residual disparity.

\subsection{Stepwise Regression Stereo}\label{SRSTEREO}

In this section, we describe the key components that implement SR-Stereo 
and show how they can be plugged into iteration-based methods. 
Using IGEV-Stereo~\cite{xu2023iterative} as the basic model, 
the overall architecture is shown in Figure~\ref{fig:3}. 
Intuitively, we retain the original feature extraction 
and cost-volume construction, but reconstruct the update unit and its 
regression objective. In addition, we propose a loss weight for the new 
update unit that is related to the regression objective scale.

\subsubsection{Feature extraction and Cost volume Construction}
We do not change the feature extraction and cost-volume construction of 
IGEV-Stereo~\cite{xu2023iterative}, but still describe them in this subsection.
% (if you are familiar with these steps, you can skip this subsection).

\textit{Feature Encoder.} 
We use MobileNetV2~\cite{sandler2018mobilenetv2} 
pre-trained on ImageNet~\cite{krizhevsky2017imagenet} 
and a series of upsampling blocks to extract multi-scale features 
$ f_{1}^{left,i}\in\mathbb{R}^{C_{i}\times H/2^{i+1} \times W/2^{i+1} }$ 
from the left image $I_{left}\in \mathbb{R} ^{3\times H\times W}$
and $ f_{1}^{right,i}\in\mathbb{R}^{C_{i}\times H/2^{i+1} \times W/2^{i+1}}$ 
from the right image $I_{right}\in \mathbb{R} ^{3\times H\times W}$, respectively.
The $i=1,2,3,4$ and the feature size $C_{i}=48, 64, 192, 160$.

\textit{Context Encoder.} 
We use a series of residual blocks~\cite{he2016deep} 
and downsampling layers to extract 
multi-scale context features 
$ f_{2}^{left,j} \in \mathbb{R}^{C_{j}\times H/2^{j+1} \times W/2^{j+1} }$ ($j$=1,2,3,4 and $C_{j}$=128)
from the left image. 
These features are inserted into the stepwise regression 
unit to provide global information.

\textit{Cost Volume.}
By combining Geometry Encoding Volume (GEV) and All-pairs Correlations (APC), 
we construct the cost volume $G_{c}$ based on $f_{1}^{left,1}$ and $f_{1}^{right,1}$. 
Then, we use a lightweight 3D UNet to aggregate the GEV and obtain 
the initial disparity $d_{init} \in \mathbb{R}^{1\times H/4 \times W/4}$.

\subsubsection{Stepwise Regression Architecture}
In this subsection, we detail the two key components 
that implement SR-Stereo: the stepwise regression 
unit and the regression objective segment.

\begin{figure}[t]
  \centering
  \includegraphics[width=0.49\textwidth]{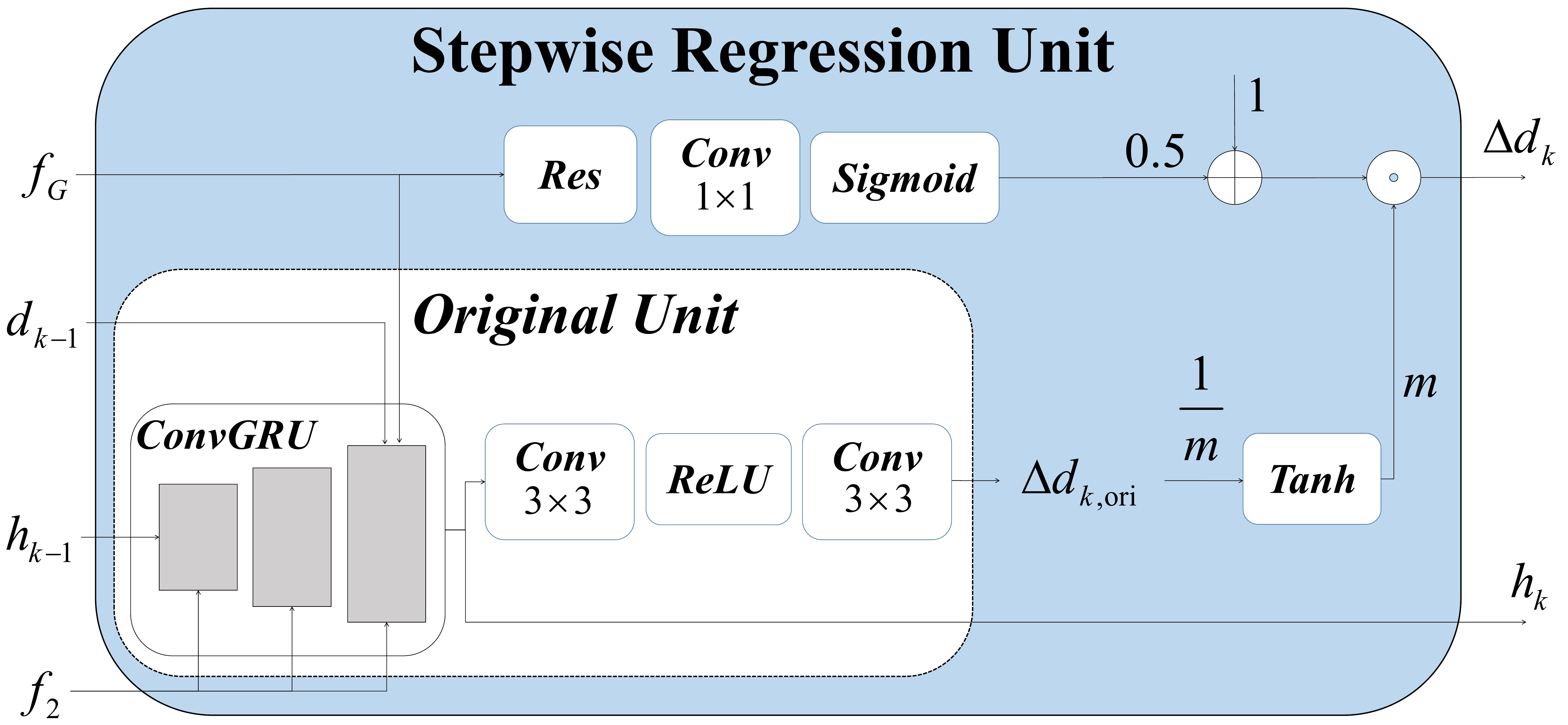}
  \caption{Architecture of the stepwise regression unit. 
  The $m$ is the hyperparameter that controls the range 
  of the output disparity clip. 
  The $\Delta d_{k,ori}$ is the residual disparity output of the original update unit,
   and $Res$ is the residual layer.}
  \label{fig:4}
\end{figure}

\textit{Stepwise Regression Unit.}
Figure~\ref{fig:4} shows the architecture of the 
stepwise regression unit, which highlights the improvements 
over the original update unit. 
For each regression, we index a set of features $f_{G}$ from 
the cost volume $G_{c}$ using the current disparity-centered 
set as follows:
\begin{equation}
  f_{G} =\sum_{r=-4}^{4} concat\left \{ G_{c}(d_{k-1}+r), G_{c}^{p}(d_{k-1}/2+r) \right \}
\end{equation}
where $d_{k-1}$ is the current disparity ($d_0$ = $d_{init}$) 
and $p$ denotes the average pooling operation.
We use these features from $G_{c}$ along with the current disparity $d_{k-1}$ to update the 
hidden state $h_{k-1}$ of ConvGRU~\cite{xu2023iterative}:
\begin{equation}
  h_{k} = ConvGRU(f_{G},d_{k-1},h_{k-1},f_{2}^{left})
\end{equation}
where $h_{k}$ is the updated hidden state.
Then, the $h_{k}$ and $f_{G}$ are 
passed into a series of convolutional layers 
and residual layers~\cite{he2016deep} to generate the disparity clip 
$\Delta d_{k}\in \mathbb{R}^{1\times H/4 \times W/4}$ as follows:
\begin{multline}
  \hfil \Delta d_{k} = tanh(\frac{\Delta d_{k,ori}}{m} ) \times m \odot (1 + 0.5w) \hfil
\end{multline}
in which
\begin{equation}
  w = \sigma (Conv_{1\times 1}(Res(f_{G})))
\end{equation}
\begin{equation}
  \Delta d_{k,ori} = Conv_{3\times 3}(Relu(Conv_{3\times 3}(h_{k})))
\end{equation}
where $\sigma$ denotes the sigmoid function, 
$Res$ is the residual layer, 
$\odot$ denotes the Hadamard product,
and $\Delta d_{k,ori}\in \mathbb{R}^{1\times H/4 \times W/4}$ 
is the residual disparity output of the original update unit. 
We set the hyperparameter $m$ to constrain the approximate 
range of the disparity clips, and use the weight map $w$ 
to adaptively adjust the constraint magnitude for different regions. 
The weight map $w$ is computed from the cost-volume feature.

Finally, we update the current disparity 
and use the Spatial Upsampling~\cite{xu2023iterative} 
to generate the full-resolution disparity
$d_{k}^{full}\in \mathbb{R}^{1\times H \times W}$ as follows:
\begin{equation}
  d_{k} = d_{k-1} + \Delta d_{k}
\end{equation}
\begin{equation}
  d_{k}^{full} = Upsampling(f_{1}^{left,i},I_{left},4d_{k})
\end{equation}

\textit{Regression Objective Segment.}
We specify how to obtain the ground truth 
disparity clip $\Delta d_{gt,k}$ and the full-resolution 
ground truth disparity $d_{gt,k}^{full}$. Specifically, the 
ground truth of the full-resolution disparity is constantly changing with the 
stepwise regression process, as shown below:
\begin{equation}
  d_{gt,k}^{full} = d_{k-1}^{full} + clip_{-6m}^{6m} (d_{gt}-d_{k-1}^{full})
\end{equation}
in which
\begin{equation}
  clip_{-M}^{M} (x) =\begin{cases}
    -M  & \text{ if } x < -M \\
    x & \text{ if } -M \leq x \leq M \\
    M & \text{ if } x > M
   \end{cases}
\end{equation}
where $d_{gt}$ is the ground truth disparity and $clip$ denotes the clipping operation. 

In addition, we use linear interpolation to generate 
the ground truth disparity clip $\Delta d_{gt,k}$:
\begin{equation}
  \Delta d_{gt,k} = clip_{-1.5m}^{1.5m} (\frac{Resize(d_{gt})}{4} -d_{k-1})
\end{equation}
where $Resize$ denotes linear interpolation downsampling.
We point out that supervision of the disparity clips is not necessary, 
but slightly improves the model's performance 
(see Section~\ref{Ablation Study} for experimental results).

\begin{figure}[t]
  \centering
  \includegraphics[width=0.49\textwidth]{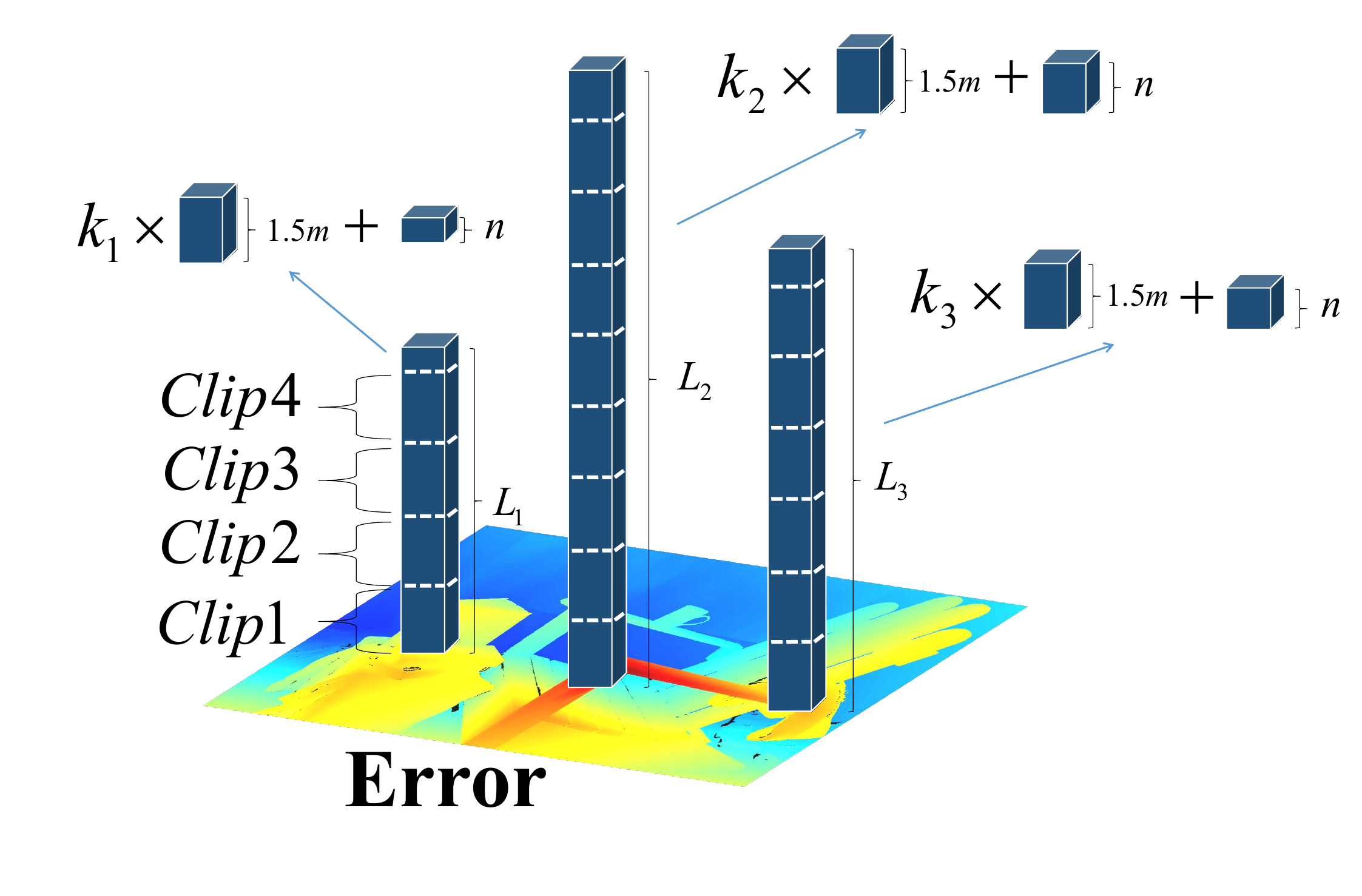}
  \caption{Long-tailed distribution of regression objectives for 
  disparity clips. SR-Stereo regresses the disparity error by predicting 
  multiple disparity clips, but the regression 
  objectives of disparity clips are long-tailed.}
  \label{fig:16}
\end{figure}

\subsubsection{Disparity Clip-Balanced Weight}
As mentioned earlier, SR-Stereo mitigates the domain discrepancies 
by predicting multiple disparity clips to regress the disparity error. 
However, this approach leads to a new problem: 
the regression objective for the disparity clips exhibits 
a long-tailed distribution, as shown in Figure~\ref{fig:16}. 
Ideally, a disparity error of size $L$ can be regressed 
by multiple clips of size $1.5m$ and a small clip of size $n$ as follows:
\begin{equation}
  L = k \times 1.5m + n
\end{equation}
where $k\in \mathbb{Z}$ and $n \in  (0,1.5m)$. 
Obviously, the distribution of regression objectives is 
long-tailed (especially when the disparity error is large), 
which limits the accuracy of small disparity clips. 
In this paper, a Disparity Clip-Balanced Weight is 
proposed to mitigate the long-tailed distribution. 
The formula for this weight is as follows:
\begin{equation}
  w_{balanced}(x) = clip_{0}^{1.5}(\left | x \right | ^{-h} ) 
\end{equation}
where $h$ is a hyperparameter that controls 
the bias towards small disparity clip. 
This weight can be flexibly inserted into existing loss functions:
\begin{equation}
  CB_{L1}(x) = w_{balanced}(x)\times \left | x \right | 
\end{equation}
\begin{equation}
  CB_{Smooth L1}(x) =\begin{cases}
    w_{balanced}(x)\times0.5x^{2}  & \text{ if } x<1 \\
    w_{balanced}(x)\times \left | x \right | & \text{ otherwise }
   \end{cases}
\end{equation}

\begin{figure*}[t]
  \centering
  \includegraphics[width=0.85\textwidth]{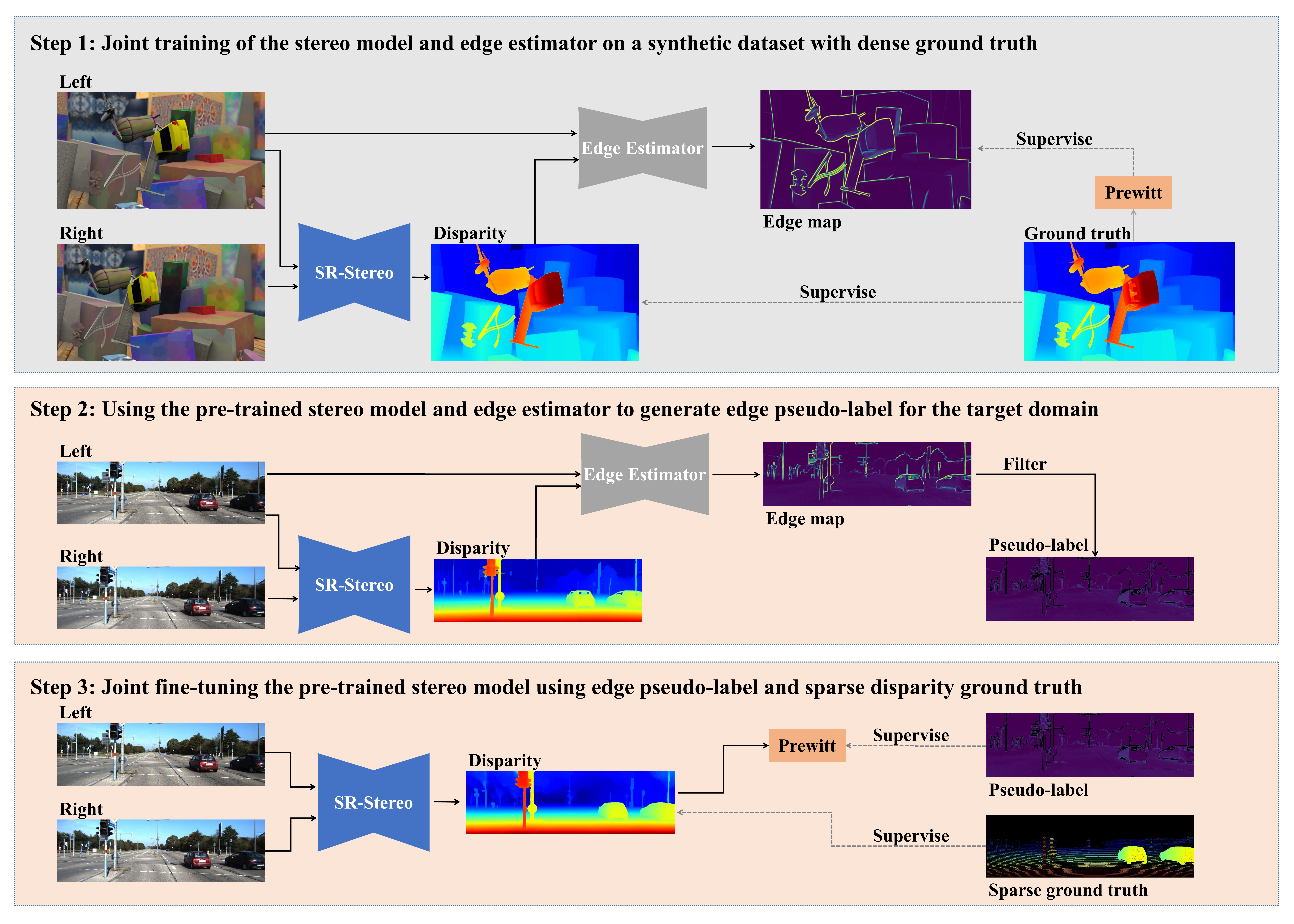}
  \caption{The overall framework of the proposed DAPE. 
  First, a robust stereo model SR-Stereo and a lightweight edge 
  estimator are pre-trained on a large synthetic dataset with 
  dense ground truth. Then, we use the pre-trained SR-Stereo and 
  edge estimator to generate the edge map of target domain, where 
  the background pixels (i.e., non-edge region pixels) are used 
  as edge pseudo-labels. Finally, we jointly fine-tune the pre-trained 
  SR-Stereo using the edge pseudo-labels and sparse ground truth disparity.}
  \label{fig:2}
\end{figure*}

\subsubsection{Loss Function}
We use Smooth $L1$ loss to supervise the initial disparity
and the disparity clips $\Delta d_k$:
\begin{equation}
  Loss_{init} = Smooth_{L1}(d_{init}^{full}-d_{gt})
\end{equation}
\begin{equation}
  Loss_{\Delta d} = \sum_{k=1}^{N} \gamma ^{N-k} CB_{Smooth L1}(\Delta d_k-\Delta d_{gt,k})
\end{equation}
where $\gamma=0.9$ and $N$ is the total number of disparity clips. We use $L1$ loss to supervise full-resolution disparities:
\begin{equation}
  Loss_{full} = \sum_{k=1}^{N} \gamma ^{N-k} CB_{L1}(d_{k}^{full}-d_{gt,k}^{full})
\end{equation}

Ultimately, the total loss function for SR-Stereo is as follows:
\begin{equation}
  \label{loss_stereo}
  Loss_{total} = Loss_{init}+Loss_{\Delta d}+Loss_{full}
\end{equation}

\subsection{Domain Adaptation based on Pre-trained Edge}\label{DAPE}
We propose a Domain Adaptation based on Pre-trained Edge (DAPE) to 
mitigate edge blurring for model fine-tuned with sparse ground truth. 
As shown in Figure~\ref{fig:2}, the proposed DAPE consists of three steps:
\begin{enumerate}
  \item{Firstly, a robust stereo model and a lightweight 
  edge estimator are pre-trained on a large synthetic 
  dataset (i.e., the source domain) with dense ground truth. 
  Specifically, the input to the edge 
  estimator is the disparity predicted by the stereo model 
  along with the RGB image.}
  \item{Then, the pre-trained stereo model 
  and the edge estimator are used to directly infer on 
  the target domain and generate corresponding edge maps. 
  To mitigate the negative impact of disparity errors in 
  reflective regions, we recommend using only the 
  relatively dense background pixels (i.e., non-edge region pixels) 
  from the edge maps as pseudo-labels 
  (following UCFNet~\cite{shen2023digging}, these pixels are defined as pseudo-labels).}
  \item{After generating the edge pseudo-labels of the target domain, 
  we use it and the sparse ground truth disparity to jointly 
  fine-tune the pre-trained stereo model. 
  % Experimental results show that the incorporation 
  % of edge pseudo-labels supervision significantly 
  % improves the disparity estimation performance 
  % of the fine-tuned model.
  }
\end{enumerate}

In the following, 
we describe the edge estimator, the edge map background 
pseudo-labels and the joint fine-tuning process in detail.

\subsubsection{Edge Estimator}\label{Edge Estimator}

For the proposed DAPE, the reliability of edge pseudo-label 
is vital to the performance of the fine-tuned model. 
Existing edge estimation methods usually take only 
an RGB image as input and use a series of complex 
2D CNNs to estimate the contour of object by extracting 
high-level features. However, this approach significantly 
increases the computational cost and limits the generalization 
performance due to the substantial differences in object categories and 
distributions across different domains.

\begin{figure}[t]
  \centering
  \includegraphics[width=0.5\textwidth]{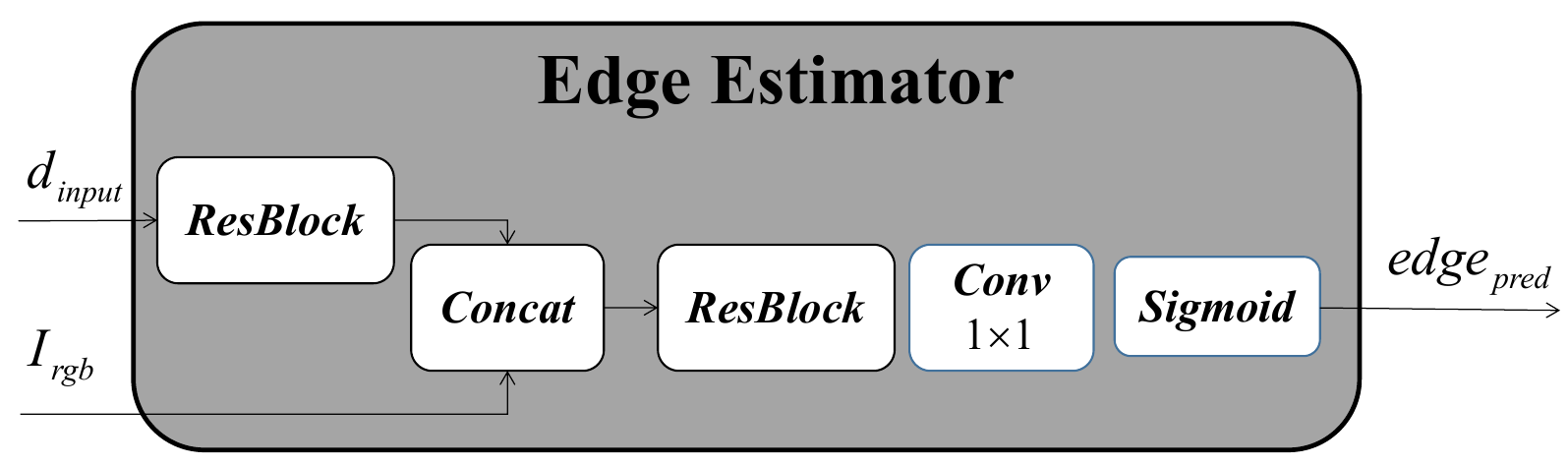}
  \caption{Architecture of the edge estimator. 
  The entire network consists of only two residual blocks and one convolution block.
  }
  \label{fig:5}
\end{figure}

In this paper, we claim that using 
disparity for edge estimation is a 
better and more robust approach because 
disparity estimation focuses on local 
geometric information, whose representations are 
similar among different domains. Moreover, 
compared to RGB image, disparity 
contains almost no object texture information, 
which greatly reduces the difficulty of 
edge estimation. Thus, a lightweight edge 
estimator is proposed to achieve accurate and 
robust edge prediction by introducing disparity. The 
architecture of the edge estimator is shown in Figure~\ref{fig:5}.
First, a residual block is used to extract the edge 
feature in the disparity:
\begin{equation}
  f_{edge} = ResBlock(d_{input})
\end{equation}
where $ResBlock$ denotes the residual block, $d_{input}$ is the disparity, 
and the number of channels for $f_{edge}$ is 29. 
In order to mitigate the potential effects of noise in the 
disparity, the RGB image is used to refine the 
edge feature map. Specifically, the edge 
feature and the corresponding RGB image are 
passed together to a residual block to achieve 
feature refinement. Finally, we use a $1\times1$ convolutional 
layer and a sigmoid layer to generate the refined edge map as follows:
\begin{equation}
  f_{refine} = ResBlock(concat(f_{edge}, I_{rgb}))
\end{equation}
\begin{equation}
  edge_{pred} = \sigma (Conv_{1\times 1}(f_{refine}))
\end{equation}
where $edge_{pred}$ is the predicted edge map, 
$I_{rgb}$ is the RGB image corresponding to the disparity, 
$\sigma$ denotes the sigmoid function, and the number of channels 
of $f_{refine}$ is 16.

\textit{Loss Function.}
We use Smooth $L1$ loss to train the proposed edge estimator on a large synthetic dataset:
\begin{equation}
  Loss_{edge} = Smooth_{L1}(edge_{pred}-edge_{gt})
\end{equation}
where $edge_{gt}$ is the ground truth edge map. 
Since the ground truth edge map is not provided in 
the dataset, we use the Prewitt operator to extract 
the edge of the ground truth disparity as $edge_{gt}$:
\begin{equation}
  edge_{gt} = \begin{cases}
    1 & \text{ if } Prewitt(d_{gt})>5 \\
    0 & \text{ otherwise }
   \end{cases}
\end{equation}
where $d_{gt}$ is the ground truth disparity and $Prewitt$ denotes 
the Prewitt operator. 

\begin{figure}[t]
  \centering
  \includegraphics[width=0.45\textwidth]{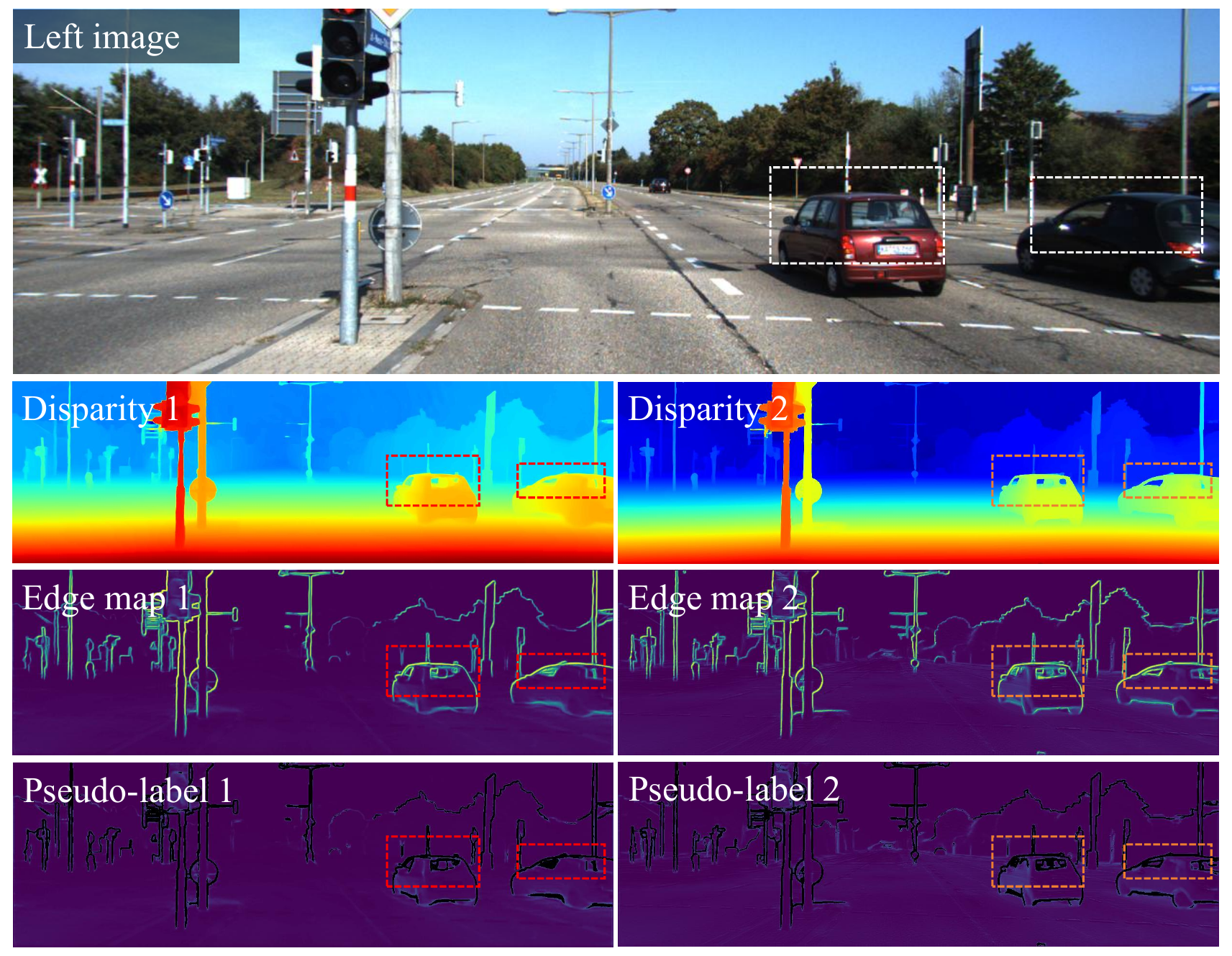}
  \caption{Generalization results of existing methods on KITTI 2015. 
  The labels 1 and 2 in the figure denote the IGEV-Stereo and the 
  proposed SR-Stereo, respectively. The edge map is represented using 
  a pseudo-color image, where the black color indicates the invalid regions. 
  As shown, the existing methods perform poorly in reflective regions, which leads to wrong edges. 
  Therefore, we propose to use only pixels in non-edge region as pseudo-label.}
  \label{fig:7}
\end{figure}

\subsubsection{Edge Map Background Pseudo-label Generation for Target Domain}\label{Edge Estimator thr}

The proposed edge estimator achieves edge 
estimation by introducing disparity, 
and therefore is affected by the accuracy 
of the disparity. As shown in Figure~\ref{fig:7}, existing 
disparity estimation methods perform badly in 
reflection regions, which leads to erroneous edges. 
In this paper, an edge map background pseudo-label 
is proposed to generate reliable supervision for fine-tuning. 
Intuitively, in an ideal edge map, 
both the edge region and non-edge region 
can be leveraged as supervision to improve 
the performance of the fine-tuned disparity estimation model. 
For instance, the disparity map typically exhibits 
high gradients in edge region, while it remains 
relatively flat in non-edge region. Erroneous edges in 
reflection regions do not affect the reliability 
of the pixels in the non-edge region. Therefore, 
we can simply utilize pixels in non-edge region as 
the pseudo-label (i.e., the edge map background pseudo-label):
\begin{equation}
  edge_{background} = \left \{ p<t\mid p\in edge_{pred} \right \}
\end{equation}
where $t$ is the threshold that controls the density of non-edge region.

\subsubsection{Joint Fine-tuning on the Target Domain}
After obtaining the edge map background pseudo-label, we 
use it and the sparse ground truth disparity from the 
target domain as supervision to jointly fine-tune 
the pre-trained stereo matching model. Specifically, 
we propose an edge-aware loss to improve the detail 
of the predicted disparity as follows:
\begin{equation}
  Loss_{edge} = Smooth_{L1}(edge_{d_{pred}}-edge_{background})
\end{equation}
where $edge_{d_{pred}}$ is the edge map corresponding to 
predicted disparity $d_{pred}$. As in Section~\ref{Edge Estimator}, we use the 
Prewitt operator to extract the edge map of the 
predicted disparity:
\begin{equation}
  edge_{d_{pred}} = \sigma(10 \times (Prewitt(d_{pred})-5))
\end{equation}
The total loss of the joint fine-tuning process is shown below:
\begin{equation}
  Loss_{DAPE} = Loss_{stereo} + Loss_{edge}
\end{equation}
where $Loss_{stereo}$ is the loss of the original 
stereo matching method, e.g., Eq~\ref{loss_stereo}.

\section{Experiments}

\subsection{Datasets and Evaluation Metrics}\label{dataset}

\textbf{SceneFlow}~\cite{mayer2016large} is a large synthetic dataset that 
consists of 35,454 training pairs and 4,370 testing pairs, 
with a resolution of $960\times540$. It provides dense ground truth 
for optical flow and stereo matching. We utilize this dataset 
to pretrain the proposed SR-Stereo and the edge estimator with 
3-pixel error and end-point error (EPE) as evaluation metrics.

\textbf{Middlebury 2014}~\cite{scharstein2014high} consists of two 
batches of indoor image pairs. The first batch provides 15 training 
pairs and 15 test pairs, with each scene provided in three different 
resolutions. The second batch provides 13 additional training pairs, 
but contains only one resolution. All training pairs are provided with 
dense hand-labeled ground truth disparity, which ranges from 0 to 300. 
We directly use the training pairs from the first batch to evaluate 
the generalization performance of SR-Stereo. For DAPE, we utilize the 
additional 13 image pairs to fine-tune the model and evaluate its 
performance using the 15 training pairs from the first batch. The 
evaluation metric used is the 2-pixel error.

\textbf{ETH3D}~\cite{schops2017multi} is a grayscale image dataset 
that includes a variety of indoor and outdoor scenes. It consists 
of 27 training pairs and 20 test pairs. The dataset provides sparsely 
labeled ground truth disparities for the training pairs, ranging 
from 0 to 60 (the smallest among several datasets). Similar to the 
usage in Middbury 2014, we employ the training pairs to directly 
assess the generalization performance of the proposed method, 
using the 1-pixel error as the evaluation metric. Additionally, 
we use 14 training pairs as fine-tuning samples and 13 training 
pairs as test samples to evaluate the effectiveness of the proposed 
Domain Adaptation based on Pre-trained Edge.

\textbf{KITTI 2012}~\cite{geiger2012we} and \textbf{KITTI 2015}~\cite{menze2015object}
are datasets of real-world driving scenes. KITTI 2012 consists of 194 training pairs 
and 195 test pairs, while KITTI 2015 contains 200 training pairs and 200 test pairs. 
Both datasets provide sparse ground truth disparity (sparsest in several datasets) 
obtained using lidar. The disparity values range from 0 to 230. 
In Section~\ref{DAPE_gen}, we merge KITTI 2012 and KITTI 2015, referred to as \textbf{KITTI}. 
For evaluation, we adopt the 3-pixel error as the metric. To assess the efficacy of our 
proposed DAPE method, we allocate 80\% of the KITTI training set for fine-tuning the model 
and reserve the remaining 20\% for validation. 

\begin{table*}[t]
  \caption{Ablation study of SR-Stereo. The baseline (i.e. *) is IGEV-Stereo. 
  The $num_{GRU}$ denotes the usage times of ConvGRU, 
  while the $num_{SRU}$ denotes the usage times of the proposed stepwise regression unit. 
  All methods run 15 disparity updates during inference ($num_{GRU} + num_{SRU} = 15$). 
  The final configuration of $m$ is underlined. \textbf{Bold}: Best. \textcolor{blue}{Blue}: Second.}
  \label{tab:as_dsr}
      \centering
      \begin{tabular}{|l|cccc|cc|c|c|c|}
          \hline
          \multirow{2}{*}{Index}  & \multicolumn{4}{c|}{Variations} & \multicolumn{2}{c|}{SceneFlow} & Middlebury-H & ETH3D & \multirow{2}{*}{Params.(M)} \\
            & $num_{GRU}$ & $num_{SRU}$ & $m$ & $Loss_{\Delta d}$ & EPE(px) & $>3px$ & $>2px(\%)$ & $>1px(\%)$ & \\
          \hline
          (a)* & 15 & 0 & - & - & 0.72 & 3.65 & 8.44 & 4.49 & 12.60 \\
          \hline
          (b) & 0 & 15 & 1 & - & 0.71 & 3.61 & 8.42 & \textcolor{blue}{4.10} & 12.77 \\
          (c) & 0 & 15 & 2 & - & \textbf{0.69} & \textcolor{blue}{3.51} & 7.89 & 4.47 & 12.77\\
          (d) & 0 & 15 & 3 & - & \textcolor{blue}{0.70} & 3.55 & 7.66 & 4.44 & 12.77 \\
          (e) & 0 & 15 & 4 & - & \textbf{0.69} & 3.54 & \textcolor{blue}{7.31} & 5.14 & 12.77 \\
          \hline
          (f) & 0 & 15 & \underline{2} & $\surd $ & \textcolor{blue}{0.70} & \textbf{3.49} & \textbf{6.78} & \textbf{4.05} & 12.77 \\
          (g) & 0 & 15 & 3 & $\surd $ & \textbf{0.69} & \textbf{3.49} & 7.44 & 4.15 & 12.77 \\
          \hline
          (h) & 4 & 11 & 2 & $\surd $ & \textbf{0.69} & \textbf{3.49} & 7.50 & 4.61 & 12.77 \\
          (i) & 4 & 11 & 3 & $\surd $ & \textcolor{blue}{0.70} & 3.56 & 7.80 & 4.61 & 12.77 \\
          (j) & 8  & 7 & 2 & $\surd $ & \textbf{0.69} & \textbf{3.49} & 7.77 & \textcolor{blue}{4.10} & 12.77 \\
          (k) & 8  & 7 & 3 & $\surd $ & \textcolor{blue}{0.70} & 3.52 & 7.91 & 4.79 & 12.77 \\
          \hline
      \end{tabular}
\end{table*}

\subsection{Implementation Details}

In this paper, we implement the proposed methods using pytorch 
and conduct experiments using two NVIDIA RTX 3090 GPUs. For all the experiments, 
we use the AdamW optimizer and one-cycle learning rate schedule, 
as well as the same data augmentation strategies. Specifically, we 
preprocess the training pairs by applying the saturation 
transform and randomly cropping them to ensure a consistent 
image size ($320\times512$ for ETH3D, $384\times1024$ for 
Middbury 2014, and $320\times672$ for other datasets). 
Below, we provide a detailed description of the training 
settings for SR-Stereo and DAPE, respectively.

\subsubsection{\textbf{SR-Stereo}}\label{SR-Stereo_final}
In the experimental section, 
following Selective-Stereo~\cite{wang2024selective} and MoCha-Stereo~\cite{chen2024mocha}, 
we implement the proposed SR-Stereo on the basis of IGEV-Stereo for a fair comparison. 
All ablation versions of SR-Stereo are trained on SceneFlow with a batch 
size of 4 for 50k steps, while the final version of SR-Stereo (represented by SR-IGEV)
is trained on SceneFlow with a batch size of 8 for 200k steps. 
The final model and ablation experiments are 
conducted using a one-cycle learning rate 
schedule with learning rates of 0.0002 and 0.0001, respectively. 
Following some generalized stereo methods~\cite{chang2023domain,zhang2022revisiting,rao2023masked}, we directly 
evaluate cross-domain performance on the training sets 
of KITTI, Middlebury, and ETH3D.

\subsubsection{\textbf{DAPE}}\label{DAPE_final}
For the experiments related to the edge estimator, we jointly 
train the stereo model and edge estimator on 
SceneFlow with a batch size of 4 for 50k steps, using a one-cycle 
learning rate schedule with a learning rate of 0.0001. Then, We use 
the pre-trained stereo model and edge estimator to generate 
edge pseudo-labels for target domains. Following existing methods~\cite{lipson2021raft, zhao2023high, li2022practical}, 
we adopt different settings of fine-tuning process for different datasets. 
For the KITTI, we adopt a batch size of 4 and fine-tune 
the model for 50k steps with an initial learning rate of 
0.0001. As for the ETH3D, we use a batch size of 2 and 
fine-tune the model for 2,000 steps, also with an initial 
learning rate of 0.0001. In the case of the Middlebury 2014, 
we utilize a batch size of 2 and fine-tune the model for 4,000 steps, 
starting with an initial learning rate of 0.00002.

\subsection{Ablation Study}\label{Ablation Study}
In this section, we explore the effectiveness and optimal configuration 
of each component of SR-Stereo.

\subsubsection{Stepwise Regression Architecture}
We explore the optimal settings for stepwise regression architecture as well as its effectiveness. 
Table~\ref{tab:as_dsr} shows the results of stepwise regression architecture in different configurations.
In the majority of configurations, the incorporation of stepwise regression architecture 
significantly enhances the performance of the baseline model across different datasets. 

In lines (b) to (e), experiments conducted with varying ranges (m) of disparity clips demonstrate 
that the choice of disparity clip range impacts the performance on different 
datasets. For datasets with a small range of disparity, smaller disparity clips are preferred, 
while datasets with a larger range of disparity benefit from larger disparity clips. 
Interestingly, the stepwise regression architecture achieves consistent and stable performance across all datasets when $m$ = 2.

In lines (f) to (g), our method achieves further performance improvement through the supervision of 
disparity clips. By incorporating supervision specifically on disparity clips, the model's 
accuracy and generalization capability are enhanced.

In lines (h) to (k), we investigate the impact of the number of stepwise regression units employed 
in the architecture. Experimental results reveal that increasing the usage times of stepwise 
regression units leads to better generalization performance of the model.

\subsubsection{Configuration of Disparity Clip-Balanced Weight}
We explore the effectiveness of Disparity Clip-Balanced Weight. 
As shown in Table~\ref{tab:as_dcb}, the utilization of Disparity Clip-Balanced Weight 
significantly improves the performance on multiple datasets. 
As mentioned previously, when the disparity error is split into multiple clips, 
the imbalance problem is shifted from the distribution 
of disparity between different domains to the distribution of disparity clips within 
the same domain. Therefore, our proposed stepwise regression architecture and Disparity Clip-Balanced Weight is 
an effective combination for achieving excellent performance.

\begin{table}[t]
  \caption{Ablation study of Disparity Clip-Balanced Weight. 
  We interpolate the proposed disparity clip-balanced weight into 
  the loss function of SR-Stereo. All methods run 15 
  disparity updates during inference. 
  The final configuration of the $h$ is underlined. \textbf{Bold}: Best.}
  \label{tab:as_dcb}
  \centering
  \begin{tabular}{|c|c|cc|c|c|}
  \hline
  \multirow{2}{*}{Methods} & \multirow{2}{*}{$h$} & \multicolumn{2}{c|}{SceneFlow} & Middlebury-F & ETH3D \\
  & & EPE(px) & $>3px$ & $>2px$ & $>1px$ \\
  \hline
  IGEV-Stereo & - & 0.72 & 3.65 & 17.47 & 4.49 \\
  \hline
  \multirow{4}{*}{SR-Stereo} & - & 0.70 & 3.49 & 14.99 & 4.05 \\
                             & 0.1 & \textbf{0.69} & 3.44 & 14.74 & 3.93 \\
  & 0.3 & \textbf{0.69} & 3.32 & 14.90 & 4.18 \\
  & \underline{0.5} & 0.70 & \textbf{3.23} & \textbf{14.23} & \textbf{3.82} \\
  \hline
  \end{tabular}
\end{table}

\begin{table}[t]
  \caption{Comparison of the efficiency of disparity update units 
  on different architectures. Compared with IGEV-Stereo, 
  the proposed SR-Stereo achieves significantly better disparity 
  estimation and cross-domain generalization with the same number
   $N$ of updates. \textbf{Bold}: Better.}
  \label{tab:as_nsr}
  \centering
  \begin{tabular}{|c|c|cc|c|c|}
  \hline
  \multirow{2}{*}{Methods} & \multirow{2}{*}{$N$} & \multicolumn{2}{c|}{SceneFlow} & Middlebury-H & ETH3D \\
  & & EPE(px) & $>3px$ & $>2px$ & $>1px$ \\
  \hline
  IGEV-Stereo & \multirow{2}{*}{9} & 0.73 & 3.69 & 8.89 & 4.74 \\
  SR-Stereo   &                    & \textbf{0.72} & \textbf{3.32} & \textbf{7.66} & \textbf{3.93} \\
  \hline
  IGEV-Stereo & \multirow{2}{*}{12} & 0.72 & 3.65 & 8.59 & 4.57 \\
  SR-Stereo   &                     & \textbf{0.71} & \textbf{3.26} & \textbf{7.31} & \textbf{3.93} \\
  \hline
  IGEV-Stereo & \multirow{2}{*}{15} & 0.72 & 3.65 & 8.44 & 4.49 \\
  SR-Stereo   &                     & \textbf{0.70} & \textbf{3.23} & \textbf{7.17} & \textbf{3.82} \\
  \hline
  IGEV-Stereo & \multirow{2}{*}{18} & 0.71 & 3.63 & 8.40 & 4.43 \\
  SR-Stereo   &                     & \textbf{0.70} & \textbf{3.22} & \textbf{7.22} & \textbf{3.86} \\
  \hline
  IGEV-Stereo & \multirow{2}{*}{21} & 0.71 & 3.63 & 8.44 & 4.42 \\
  SR-Stereo   &                     & \textbf{0.70} & \textbf{3.22} & \textbf{7.15} & \textbf{3.81} \\
  \hline
  IGEV-Stereo & \multirow{2}{*}{32} & 0.73 & 3.68 & 8.18 & 4.44 \\
  SR-Stereo   &                     & \textbf{0.71} & \textbf{3.24} & \textbf{7.14} & \textbf{3.85} \\
  \hline
  \end{tabular}
\end{table}

\subsubsection{Number of Stepwise Regression Units}
In the practice of this paper, SR-Stereo remains 
essentially an iteration-based method. Therefore, our SR-Stereo can trade 
off efficiency and performance by adjusting the number of update 
units. As shown in Table~\ref{tab:as_nsr}, SR-Stereo can achieve better performance 
for the same number of updates compared to the baseline, IGEV-Stereo. 
This result also shows that the range constraint on the 
update units does not reduce the convergence speed of the disparity, 
but rather makes the updated disparity more accurate.

% \subsection{Easy Generalization of the Stepwise Regression Architecture}\label{Extension}
\subsubsection{The Simplified Stepwise Regression Architecture}\label{Extension}
Notably, the proposed stepwise regression architecture contains three key components: 
the Stepwise Regression Unit, Regression Objective Segment (ROS), and 
Disparity Clip-Balanced Weight (DCB). Except for the Stepwise Regression Unit, 
the other two do not involve changes in the network structure. To further 
demonstrate the plug-and-play of the architecture, we apply the 
proposed ROS and DCB to the iteration-based methods RAFT-Stereo and IGEV-Stereo.

\begin{table}[t]
  \caption{
    Extension results for the simplified stepwise regression architecture. 
    For inference, IGEV-Stereo runs 15 disparity updates, while 
    RAFT-Stereo runs 32 disparity updates. ROS:Regression Objective Segment. 
    DCB: Disparity Clip-Balanced Weight ($h$=0.5). 
    Gray: performance is improved after using the proposed method. \textbf{Bold}: Best.}
  \label{tab:as_esr}
  \centering
  \begin{tabular}{|c|c|ccc|}
  \hline
  \multirow{2}{*}{Methods} & \multirow{2}{*}{$m$} & SceneFlow & Middlebury-H & ETH3D \\
  & & $>3px(\%)$ & $>2px(\%)$ & $>1px(\%)$ \\
  \hline
  IGEV-Stereo & -  & 3.65 & 8.44 & 4.49 \\
  \hline
  \multirow{3}{*}{IGEV.+ROS} & 2 & \cellcolor{gray!20}3.50 & \cellcolor{gray!20}8.09 & \cellcolor{gray!20}4.40 \\
                             & 3 & \cellcolor{gray!20}3.48 & \cellcolor{gray!20}8.06 & 4.71 \\
                             & 4 & \cellcolor{gray!20}3.48 & \cellcolor{gray!20}8.10 & \cellcolor{gray!20}4.37 \\
  \hline
  \multirow{3}{*}{IGEV.+ROS+DCB} & 2 & \cellcolor{gray!20}3.32 & \cellcolor{gray!20}7.68 & \cellcolor{gray!20}3.87 \\
                             & 3 & \cellcolor{gray!20}3.26 & \cellcolor{gray!20}\textbf{7.61} & \cellcolor{gray!20}4.22 \\
                             & 4 & \cellcolor{gray!20}\textbf{3.24} & \cellcolor{gray!20}7.78 & \cellcolor{gray!20}\textbf{3.47} \\
  \hline
  RAFT-Stereo & - & 3.96 & 14.71 & 4.51 \\
  \hline
  \multirow{3}{*}{RAFT.+ROS} & 3 & 4.25 & 15.78 & \cellcolor{gray!20}3.86 \\
                             & 4 & 4.01 & \cellcolor{gray!20}14.42 & \cellcolor{gray!20}4.02 \\
                             & 5 & \cellcolor{gray!20}3.76 & \cellcolor{gray!20}12.26 & \cellcolor{gray!20}4.13 \\
  \hline
  \multirow{3}{*}{RAFT.+ROS+DCB} & 3 & \cellcolor{gray!20}3.70 & \cellcolor{gray!20}13.97 & \cellcolor{gray!20}3.60 \\
                             & 4 & \cellcolor{gray!20}3.56 & \cellcolor{gray!20}12.64 & \cellcolor{gray!20}3.63 \\
                             & 5 & \cellcolor{gray!20}\textbf{3.42} & \cellcolor{gray!20}\textbf{11.88} & \cellcolor{gray!20}\textbf{3.59} \\
  \hline
  \end{tabular}
\end{table}

\begin{table*}[t]
  \caption{Quantitative evaluation on SceneFlow and KITTI. 
  Our SR-IGEV run 16 disparity updates during inference. 
  \textbf{Bold}: Best. \underline{Underline}: Second. \textcolor{blue}{Blue}: Third.
  *: the baseline. \textcolor{red}{Bottom right corner}: Percentage improvement compared to the baseline.}
  \label{tab:Benchmarks}
  \centering
  \begin{tabular}{|l|c|c|ccc|ccccc|c|}
      \hline
      \multirow{2}{*}{Methods} & \multirow{2}{*}{Years} & \multirow{2}{*}{SceneFlow} & \multicolumn{3}{c|}{KITTI 2015} & \multicolumn{5}{c|}{KITTI 2012} & KITTI \\
        & & & D1-bg & D1-fg & D1-all & 2-noc & 2-all & 3-noc & 3-all & EPE-all & Time(s) \\
        \hline
      PSMNet~\cite{chang2018pyramid} & 2018 & 1.09 & 1.86 & 4.62 & 2.32 & 2.44 & 3.01 & 1.49 & 1.89 & 0.6 & 0.41 \\
      GANet~\cite{zhang2019ga} & 2019 & 0.80 & 1.48 & 3.46 & 1.81 & 1.89 & 2.50 & 1.19 & 1.60  & 0.5 & 1.80 \\
      GwcNet~\cite{guo2019group} & 2019 & 0.98 & 1.74 & 3.93 & 2.11 & 2.16 & 2.71 & 1.32 & 1.70  & 0.5 & 0.32 \\
      AcfNet~\cite{zhang2020adaptive} & 2020 & 0.87 & 1.51 & 3.80 & 1.89 & 1.83 & 2.35 & 1.17 & 1.54  & 0.5 & 0.48 \\
      RAFT-Stereo~\cite{lipson2021raft} & 2021 & 0.56 & 1.58 & 3.05 & 1.82 & 1.92 & 2.42 & 1.30 & 1.66  & 0.5 & 0.38 \\
      CREStereo~\cite{li2022practical} & 2022 & - & 1.45 & 2.86 & 1.69 & 1.72 & 2.18 & 1.14 & 1.46  & 0.5 & 0.41\\
      ACVNet~\cite{xu2022attention} & 2022 & 0.48 & 1.37 & 3.07 & 1.65 & 1.83 & 2.35 & 1.13 & 1.47  & 0.5 & 0.20 \\
      DLNR~\cite{zhao2023high} & 2023& 0.48 & 1.37 & \textcolor{blue}{2.59} & 1.76 & - & - & - & -  & - & 0.28 \\
      Croco-Stereo~\cite{weinzaepfel2023croco} & 2023 & - & 1.38 & 2.65 & 1.59 & - & - & - & -  & - & 0.93\\
      UPFNet~\cite{shen2023digging} & 2023 & - & 1.38 & 2.85 & 1.62 & 1.67 & \textcolor{blue}{2.17} & 1.09 & 1.45  & 0.5 & 0.25 \\
      IGEV-Stereo*~\cite{xu2023iterative} & 2023 & 0.47 & 1.38 & 2.67 & 1.59 & 1.71 & \textcolor{blue}{2.17} & 1.12 & 1.44  & \textbf{0.4} & \underline{0.18} \\
      NMRF~\cite{guan2024neural} & 2024 & \textcolor{blue}{0.45} & \textbf{1.28} & 3.13 & 1.59 & \textbf{1.59} & \underline{2.07} & \textbf{1.01} & \textbf{1.35}  & - & \textbf{0.09} \\
      ADL-GwcNet~\cite{xu2024adaptive} & 2024 & 0.62 & 1.42 & 3.01 & 1.68 & \underline{1.65} & \textcolor{blue}{2.17} & \underline{1.05} & 1.42  & - & - \\
      DKT-IGEV~\cite{zhang2024robust} & 2024 & - & 1.46 & 3.05 & 1.72 & - & - & 1.22 & 1.56  & - & - \\
      MoCha-Stereo~\cite{chen2024mocha} & 2024 & \textbf{0.41} & \textcolor{blue}{1.36} & \textbf{2.43} & \textbf{1.53} & - & - & \textcolor{blue}{1.06} & \underline{1.36}  & \textbf{0.4} & 0.34 \\
      Selective-IGEV~\cite{wang2024selective} & 2024 & \underline{0.44} & \underline{1.33} & 2.61 & \underline{1.55} & \textbf{1.59} & \textbf{2.05} & 1.07 & 1.38  & \textbf{0.4} & 0.24 \\
      \hline
      SR-IGEV(Ours) & - & $\textcolor{blue}{0.45}_{\textcolor{red}{-4.3\%}}$ & 1.37 
      & $\underline{2.49}_{\textcolor{red}{-6.7\%}}$ & $\textcolor{blue}{1.56}_{\textcolor{red}{}}$ 
      & $\textcolor{blue}{1.66}_{\textcolor{red}{}}$ & $\underline{2.07}_{\textcolor{red}{-4.6\%}}$ 
      & $1.09_{\textcolor{red}{}}$ & $\underline{1.36}_{\textcolor{red}{-5.6\%}}$ & \textbf{0.4} & $\textcolor{blue}{0.19}_{\textcolor{red}{}}$ \\
      \hline
  \end{tabular}
\end{table*}

\begin{figure}[t]
  \centering
  \includegraphics[width=0.49\textwidth]{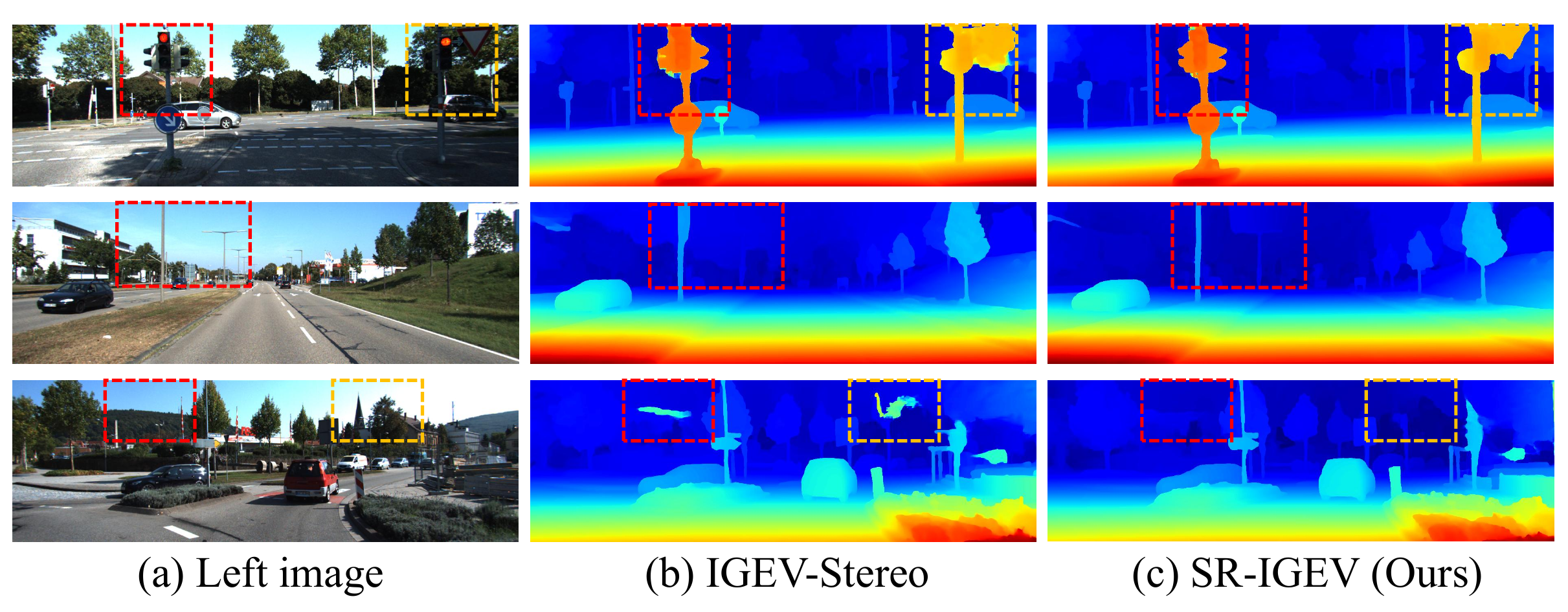}
  \caption{Qualitative results on the test set of KITTI 2015.
  Both methods run 32 updates at inference. Our SR-IGEV is more accurate for edge regions and backgrounds.
  }
  \label{fig:10}
\end{figure}

As shown in Table~\ref{tab:as_esr}, the ROS and DCB can improve 
the performance of existing iteration-based methods without changing 
the network structure, and therefore can be generalized very easily. 
In addition, we can adjust the disparity clip 
range $m$ to achieve better results according to the characteristics 
of the existing methods. For instance, in the case of RAFT-Stereo, 
where the initial disparity is set to 0, a large clip range is required 
to accelerate the disparity convergence. Therefore, the best results are 
obtained with a clip range of $m$=5. Conversely, IGEV-Stereo obtains a 
coarse initial disparity through 3D-CNN, making a smaller clip range 
more suitable to achieve stable performance improvement.

\subsection{Comparison with SOTA Methods}
\subsubsection{Benchmark Results}
In this section, we compare SR-Stereo with 
the state-of-the-art methods published on SceneFlow and KITTI.
As described in Section~\ref{SR-Stereo_final}, 
we train the SR-IGEV on SceneFlow for 200k steps. 
Then, we fine-tune the SR-IGEV on the KITTI for 50k steps.
Tables~\ref{tab:Benchmarks} shows the quantitative results. 
% We state again that the motivation of our method is to 
% improve the generalization performance of existing methods. 
As shown in Table~\ref{tab:Benchmarks}, unlike the state-of-the-art 
generalized stereo method DKT-IGEV~\cite{zhang2024robust}, our method does not deteriorate 
the in-domain performance of the baseline model. On the contrary, 
surprisingly, with a similar training strategy, our SR-IGEV achieves 
similar performance to the state-of-the-art method Selective-IGEV~\cite{wang2024selective}. More 
importantly, compared to the baseline IGEV-Stereo, our method only 
increases the time cost by 0.01s, which is far better than the 
two IGEV-based SOTA methods (Selective-IGEV~\cite{wang2024selective} and MoCha-Stereo~\cite{chen2024mocha}). 
All these results effectively demonstrate the advantages of our method.
Figure~\ref{fig:10} shows a comparison of the qualitative results of SR-Stereo 
and IGEV-Stereo on KITTI 2015. Our method is more accurate for edge regions and backgrounds.

\begin{figure}[t]
  \centering
  \includegraphics[width=0.49\textwidth]{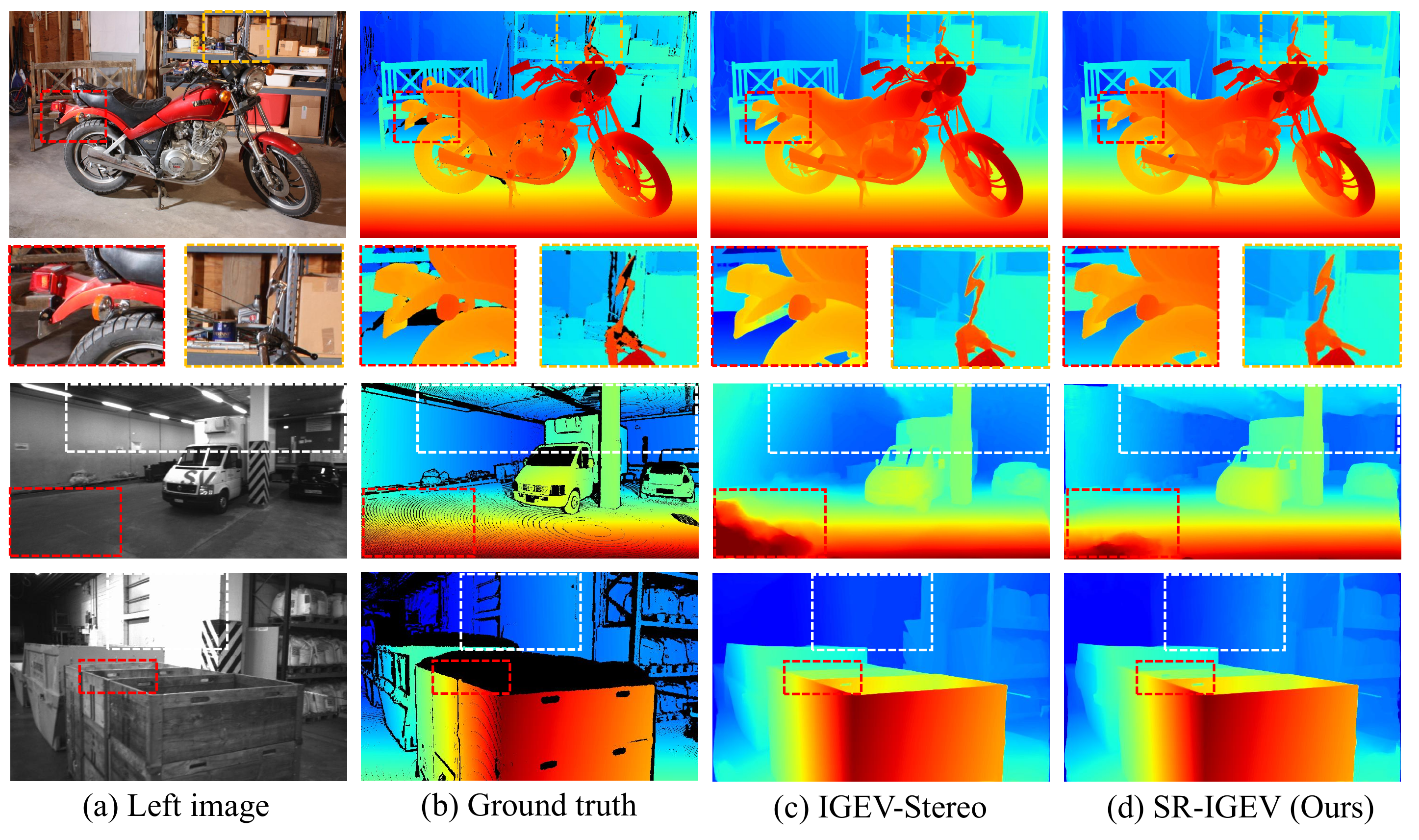}
  \caption{Generalization results on Middlebury 2014 and ETH3D. 
  All methods run 32 disparity updates during inference. 
  Our SR-IGEV performs better in the background and detail regions.}
  \label{fig:9}
\end{figure}

\subsubsection{Zero-shot Generalization}
We test the SR-IGEV directly on Middlebury, ETH3D and KITTI.
As shown in Table~\ref{tab:zero}, Our method achieves very competitive generalization performance. 
Compared with the baseline IGEV-Stereo, our method achieves an overall improvement. 
Moreover, compared to the majority of generalized stereo methods, our method achieves 
the best performance on Middlebury and ETH3D. The generalization performance of our 
method on KITTI is also very competitive among the methods of the last two years. 
We also point out that we recommend evaluating the generalization performance more 
objectively based on the results on Middlebury and ETH3D, 
as is done in IGEV-Stereo~\cite{xu2023iterative} and MoCha-Stereo~\cite{chen2024mocha}. 
Because the ground truth in KITTI is too sparse, which ignores errors in some regions.
Finally, we visualize the generalization results of SR-IGEV in Figure~\ref{fig:9}. 
It can be seen that our method performs better in the background and detail regions.

\begin{table*}[t]
  \small
  \centering
  \caption{Cross-domain generalization evaluation. 
  Our SR-IGEV run 32 disparity updates during inference.
  We pre-train our model on SceneFlow and test it directly on Middlebury 2014, ETH3D and KITTI.
  The 2-pixel error rate is used for Middlebury 2014, 
  1-pixel error rate for ETH3D and D1-error for KITTI. 
  \textbf{Bold}: Best. \underline{Underline}: Second. 
  *: the baseline. \textcolor{red}{Bottom right corner}: Percentage improvement compared to the baseline.}
  \label{tab:zero}
  \begin{tabular}{|l|c|c|ccccc|}
    \hline
      \multirow{2}{*}{Methods} & \multirow{2}{*}{Type} & \multirow{2}{*}{Years} & \multicolumn{2}{c}{Middlebury} & \multirow{2}{*}{ETH3D} & \multirow{2}{*}{KITTI-12} & \multirow{2}{*}{KITTI-15}\\
       & & & Half & Quarter & & & \\
      \hline
      PSMNet~\cite{chang2018pyramid} & & 2018 & 15.8 & 9.8 & 10.2 & 15.1 & 16.3 \\
      GANet~\cite{zhang2019ga} & & 2019 & 13.5 & 8.5 & 6.5 & 10.1 & 11.7 \\ 
      DSMNet~\cite{zhang2020domain} & & 2020 & 13.8 & 8.1 & 6.2 & 6.2 & 6.5 \\ 
      STTR~\cite{li2021revisiting} &  & 2021 & 15.5 & 9.7 & 17.2 & - & - \\
      % CFNet~\cite{shen2021cfnet} & \multirow{2}{*}{Normal} & 2021 & 15.3 & 9.8 & 5.8 & 4.7 & 5.8 \\
      RAFT-Stereo~\cite{lipson2021raft} & Normal & 2021 & 8.7 & 7.3 & 3.2 & 5.7 & 5.5 \\
      IGEV-Stereo*~\cite{xu2023iterative} & & 2023 & 7.1 & 6.2 & 3.6 & 5.2 & 6.0 \\
      MoCha-Stereo~\cite{chen2024mocha} & & 2024 & \underline{6.2} & \textbf{4.9} & 3.2 & - & - \\
      Selective-IGEV~\cite{wang2024selective} & & 2024 & 6.8 & 6.6 & 5.4 & 5.6 & 6.0 \\
      NMRF~\cite{guan2024neural} & & 2024 & - & 7.5 & 3.8 & 4.2 & 5.1 \\
      \hline
      % Graft-PSMNet~\cite{liu2022graftnet} & & 2022 & - & - & 7.7 & 4.8 & \underline{4.3} \\
      Graft-GANNet~\cite{liu2022graftnet} & & 2022 & - & - & 6.2 & 4.9 & \textbf{4.2} \\
      FC-DSMNet~\cite{zhang2022revisiting} & & 2022 & 12.0 & 7.8 & 6.0 & 5.5 & 6.2 \\ 
      HVT-GwcNet~\cite{chang2023domain} & \multirow{2}{*}{Generalized} & 2023 & 10.3 & - & 5.9 & \underline{3.9} & 5.0 \\
      Mask-LacGwcNet~\cite{rao2023masked} & & 2023 & 16.9 & - & 5.3 & 5.7 & 5.6 \\
      ADL-GwcNet~\cite{xu2024adaptive} & & 2024 & 9.1 & - & 3.8 & 4.5 & \textbf{4.2} \\
      DKT-IGEV~\cite{zhang2024robust} & & 2024 & 6.9 & - & \underline{3.1} & \textbf{3.8} & \underline{4.8} \\
      \hline
      SR-IGEV(Ours) & - & - & $\textbf{6.0}_{\textcolor{red}{-15.5\%}}$ & $\underline{6.0}_{\textcolor{red}{-3.2\%}}$ & $\textbf{3.0}_{\textcolor{red}{-16.7\%}}$ & $4.9_{\textcolor{red}{-5.8\%}}$ & $5.7_{\textcolor{red}{-5\%}}$ \\
      \hline
  \end{tabular}
\end{table*}

\begin{figure}[t]
  \centering
  \includegraphics[width=0.47\textwidth]{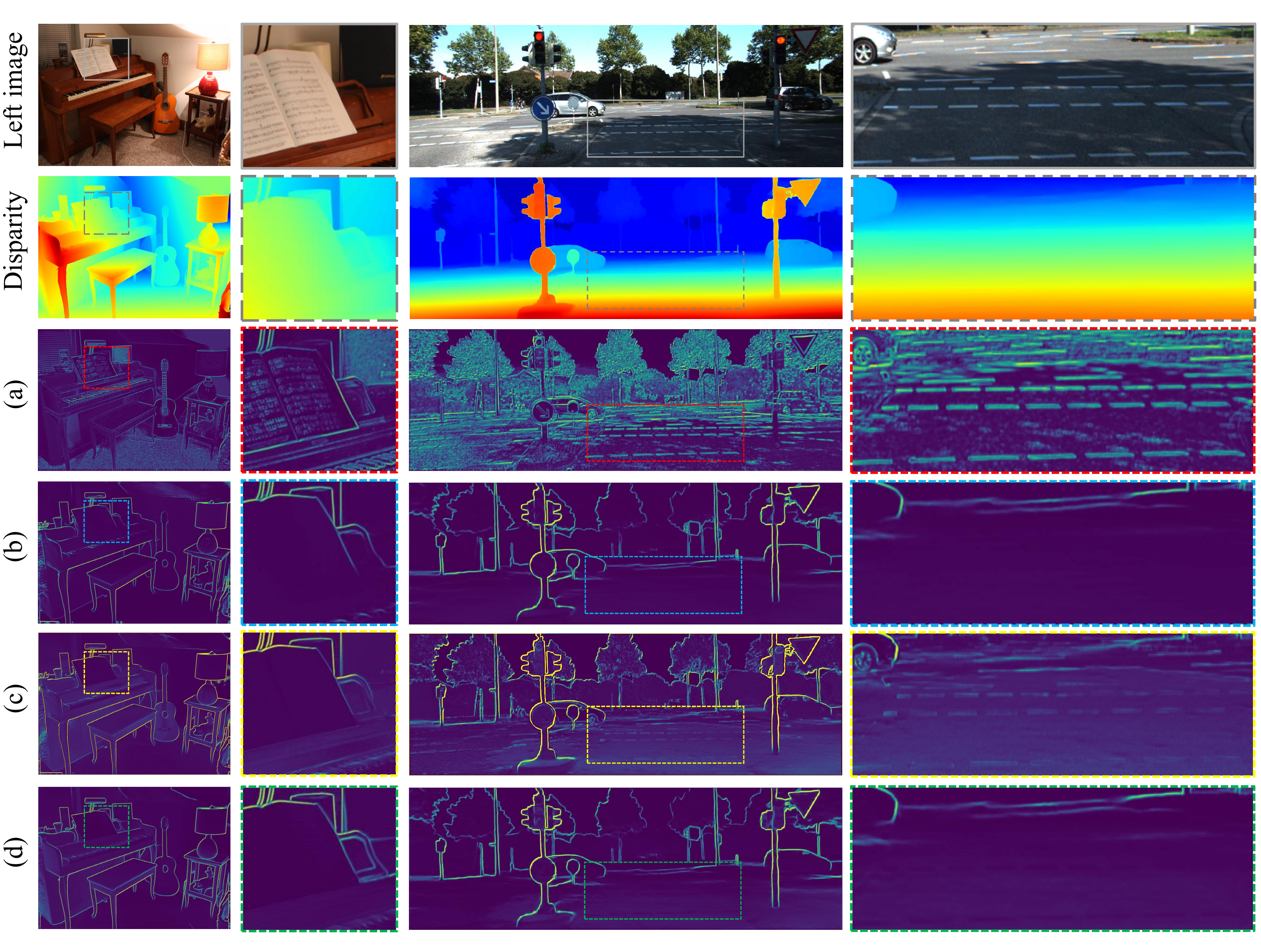}
  \caption{Edge Estimation Results for Different Inputs. 
  (a) Based on RGB image only. 
  (b) Based on disparity only. 
  (c) Based on concatenation of RGB image and disparity. 
  (d) Step-by-step input of RGB image and disparity (Figure~\ref{fig:5}).
  All methods are trained on the SceneFlow using the same 
  network architecture (i.e., 
  two residual blocks in Figure~\ref{fig:5}). Test samples 
  are obtained from Middbury 2014 and KITTI-2015.}
  \label{fig:8}
\end{figure}

\subsection{Domain Adaptation based on Pre-trained Edge}
As mentioned in Section~\ref{Overview}, in addition to proposing a robust network 
SR-Stereo, we also propose the DAPE to enhance the performance of existing models 
fine-tuned with sparse ground truth. In this section, we comprehensively 
evaluate the individual steps of DAPE and demonstrate its effectiveness 
through experimental results on KITTI, Middbury and ETH3D.

\subsubsection{Different Inputs to the Edge Estimator}\label{Edge_Estimator}
We explore the impact of different inputs on performance while 
utilizing the same edge estimator structure (Figure~\ref{fig:5}). 
Figure~\ref{fig:8} shows the qualitative results 
that highlight the differences between predicted edge maps 
obtained from different inputs. 
When only RGB images are used, the edge maps 
are excessively noisy due to the interference of 
object surface textures. In contrast, with the 
introduction of disparity, the noise in the edge map is 
significantly reduced. To further demonstrate the 
impact of edge detection on DAPE, we fine-tune 
IGEV-Stereo using different edge results. As shown in Table~\ref{tab:dape_edge}, 
the edge maps based only on RGB images worsen the effect 
of the DAPE architecture, which is particularly obvious 
on ETH3D. In addition, the three disparity-based edge 
maps do not show a significant difference in the DAPE 
architecture, and all of them improve the fine-tuning 
performance. However, considering the visualization in 
Figure~\ref{fig:8} and the diversity of signal sources, we 
recommend using the approach in Figure~\ref{fig:5}.

\begin{table}[t]
  \caption{DAPE fine-tuning results based on different edge maps. 
  The baseline is IGEV-Stereo and is fine-tuned on the sparsified 
  ground truth (see Figure~\ref{fig:11} for more details). 
  ED: Edge Detection. $\dagger$: Based on concatenation of RGB image and disparity.
  $*$: Step-by-step input of RGB image and disparity (Figure~\ref{fig:5}).
  The $t = 0.25$ in DAPE. \textbf{Bold}: Best.}
  \label{tab:dape_edge}
  \centering
  \begin{tabular}{|c|cc|ccc|}
  \hline
  \multirow{2}{*}{Methods} & \multicolumn{2}{c|}{ED Input} & \multicolumn{2}{c}{Middlebury} & \multirow{2}{*}{ETH3D} \\
  & RGB & Disparity & Full & Half & \\
  \hline
  Baseline & - & - & 11.72 & 5.14 & 1.65 \\
  \hline
  \multirow{4}{*}{DAPE} & $\surd $ & - & 11.76 & 5.01 & 2.05 \\
  & - & $\surd $ & 11.69 & \textbf{4.89} & 1.62 \\
  & $\surd \dagger $ & $\surd \dagger $ & \textbf{11.67} & 4.92 & 1.62 \\
  & $\surd * $ & $\surd * $ & 11.69 & 4.93 & \textbf{1.61} \\
  \hline
  \end{tabular}
\end{table}

\begin{figure}[t]
  \centering
  \includegraphics[width=0.49\textwidth]{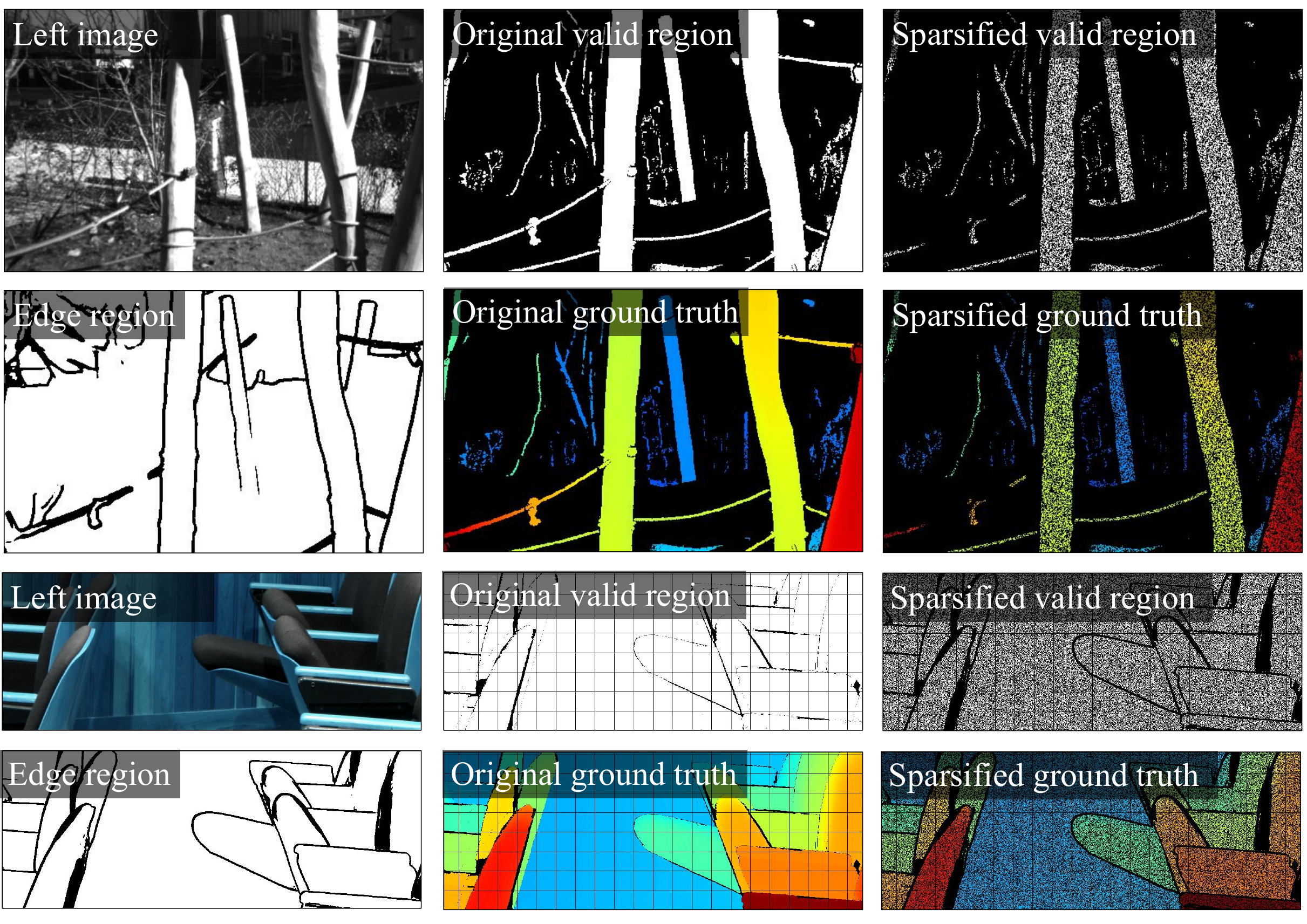}
  \caption{Visualization of ground truth sparse process for ETH3D and Middbury 2014. 
  First, we predict the edge map using the edge estimator 
  proposed in Section~\ref{Edge_Estimator}. Then, we remove the pixels in 
  the edge region of the ground truth and randomly remove 
  the pixels in the non-edge region with a probability of 0.5.}
  \label{fig:11}
\end{figure}

\subsubsection{Threshold for Edge Pseudo-label Generation}\label{DAPE_gen}

As mentioned in Section~\ref{Edge Estimator thr}, erroneous disparity estimation in ill-posed 
regions can lead to incorrect edge maps. We propose the use of threshold-based 
filtering to select pixels in non-edge region as pseudo-label. 
In this section, our focus is on exploring the threshold settings 
in different target domains. By doing so, we aim to analyze the impact 
of pseudo-label density on the adaptation of various domains. This 
analysis allows us to gain insights into the relationship between 
pseudo-label density and the effectiveness of DAPE.

\begin{table}[t]
  \caption{Domain adaptation evaluation of DAPE on ETH3D. 
  We experiment with the proposed DAPE on three models. 
  All methods are fine-tuned on the sparsified ground 
  truth and evaluated on the original ground truth. All 
  methods run 15 disparity updates during inference. 
  Gray: performance is improved after using DAPE. \textbf{Bold}: Best.}
  \label{tab:dape_eth3d}
  \centering
  \begin{tabular}{|c|l|cccc|}
  \hline
  Methods & DAPE & $>1px$ & $>0.75px$ & $>0.25px$ & EPE(px)  \\
  \hline
  \multirow{5}{*}{RAFT-Stereo} & - & 3.46 & 4.47 & 20.74 & 0.315 \\
  & $t$=0.25 & \cellcolor{gray!20}3.14 & \cellcolor{gray!20}4.29 & \cellcolor{gray!20}20.65 & \cellcolor{gray!20}0.257 \\
  & $t$=0.5 & \cellcolor{gray!20}\textbf{2.94} & \cellcolor{gray!20}\textbf{4.18} & 20.79 & \cellcolor{gray!20}0.256 \\
  & $t$=0.75 & \cellcolor{gray!20}3.21 & \cellcolor{gray!20}4.32 & 20.76 & \cellcolor{gray!20}\textbf{0.252} \\
  & $t$=1.0 & \cellcolor{gray!20}2.98 & \cellcolor{gray!20}4.26 & \cellcolor{gray!20}\textbf{20.59} & \cellcolor{gray!20}0.255 \\
  \hline
  \multirow{5}{*}{IGEV-Stereo} & - & 1.65 & 2.13 & 14.74 & 0.186 \\
  & $t$=0.25 & \cellcolor{gray!20}\textbf{1.61} & \cellcolor{gray!20}\textbf{2.12} & \cellcolor{gray!20}\textbf{14.65} & \cellcolor{gray!20}0.182 \\
  & $t$=0.5 & 1.67 & 2.20 & \cellcolor{gray!20}14.72 & \cellcolor{gray!20}\textbf{0.178} \\
  & $t$=0.75 & 1.68 & 2.18 & 14.90 & \cellcolor{gray!20}\textbf{0.178} \\
  & $t$=1.0 & 1.68 & 2.18 & 14.97 & \cellcolor{gray!20}0.179 \\
  \hline
  \multirow{5}{*}{SR-Stereo} & - & 1.59 & 1.99 & 14.31 & 0.181  \\
  & $t$=0.25 & \cellcolor{gray!20}\textbf{1.52} & \cellcolor{gray!20}\textbf{1.97} & \cellcolor{gray!20}14.23 & \cellcolor{gray!20}0.177 \\
  & $t$=0.5 & 1.64 & 2.08 & \cellcolor{gray!20}13.76 & \cellcolor{gray!20}0.178 \\
  & $t$=0.75 & 1.66 & 2.09 & \cellcolor{gray!20}\textbf{13.74} & \cellcolor{gray!20}\textbf{0.174} \\
  & $t$=1.0 & 1.69 & 2.13 & \cellcolor{gray!20}13.76 & \cellcolor{gray!20}0.176 \\
  \hline
  \end{tabular}
\end{table}

\textit{Results on ETH3D}: As mentioned in Section~\ref{dataset}, we divide the ETH3D 
training pairs into two parts which are used for fine-tuning and evaluation 
respectively. To simulate the fine-tuning process on sparse ground truth, 
the ground truth of the part used for fine-tuning is randomly removed. 
Specifically, we remove the pixels in the edge region and randomly 
remove the pixels in the non-edge region with a probability of 0.5, as shown in Figure~\ref{fig:11}.

\begin{figure}[t]
  \centering
  \includegraphics[width=0.49\textwidth]{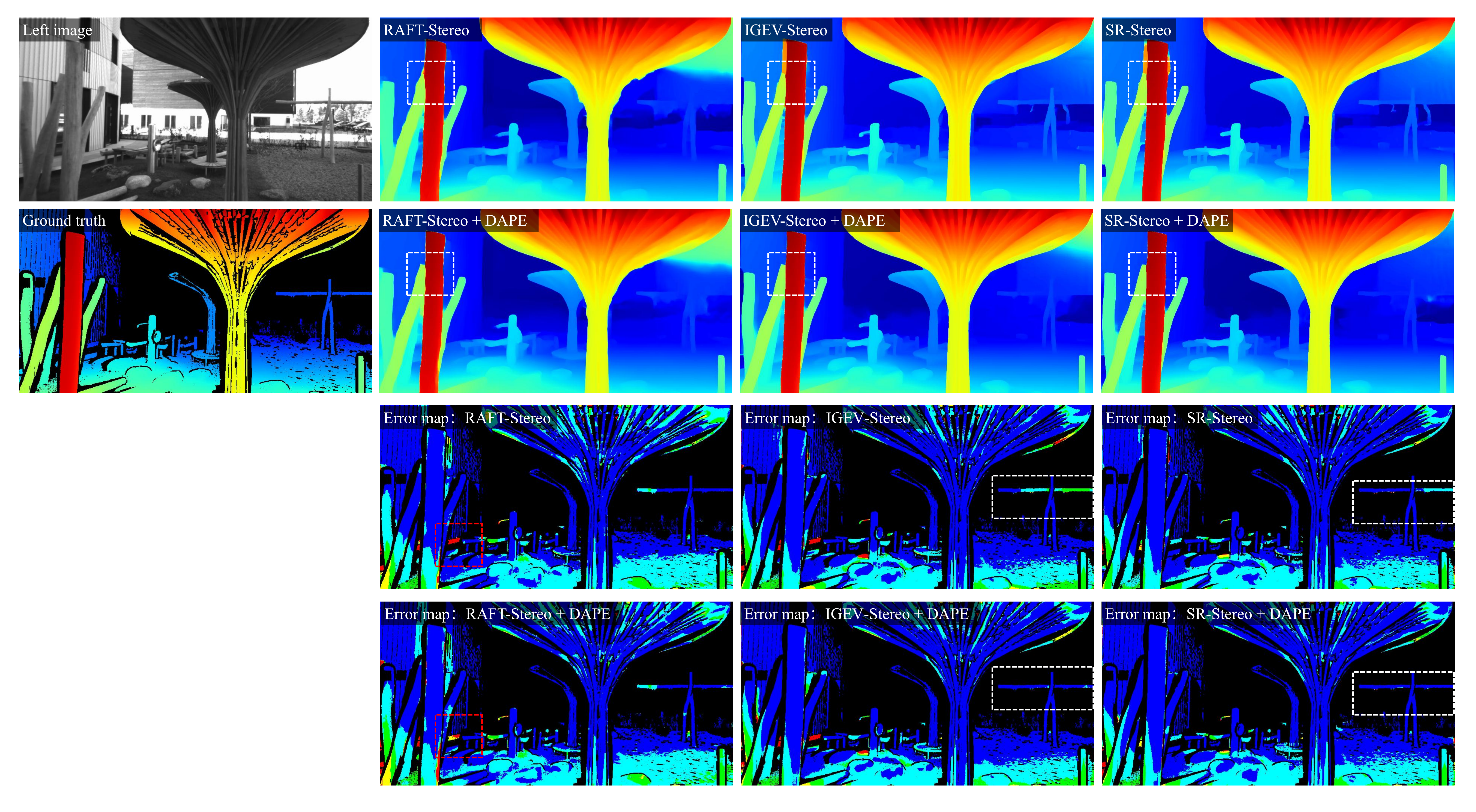}
  \caption{Qualitative disparity estimation results of DAPE on ETH3D. 
  All methods run 15 disparity updates during inference. 
  The threshold $t$ in DAPE is 0.25. In the error maps, 
  red represents a larger error, while dark blue indicates a smaller error.}
  \label{fig:12}
\end{figure}

We utilize sparsified ground truth for model fine-tuning and 
the original ground truth for evaluation, ensuring a more 
accurate assessment of the model's domain-adaptive performance.
The experimental results are presented in Table~\ref{tab:dape_eth3d}. 
As the threshold t decreases in DAPE, the density of generated 
edge pseudo-labels decreases as well. When $t$ is set to 0.25, 
all models exhibit performance improvement. This is due to the 
presence of reflective regions in some image pairs from ETH3D, 
which can lead to misleading edge map predictions. By using a 
smaller threshold, false edges are filtered out, resulting 
in more stable performance improvement. Figure~\ref{fig:12} shows the 
qualitative results of DAPE for different models. It can be 
observed that our proposed DAPE effectively improves the 
performance of the model in the detail region.

\begin{table}[!t]
  \caption{Domain adaptation evaluation of DAPE on KITTI. 
  We experiment with the proposed DAPE on three models. 
  All methods run 15 disparity updates during inference.
  Gray: performance is improved after using DAPE. \textbf{Bold}: Best.}
  \label{tab:dape_kitti}
  \centering
  \begin{tabular}{|c|l|cccc|}
  \hline
  Methods & DAPE & 3-noc & 3-all & EPE-noc & EPE-all  \\
  \hline
  \multirow{5}{*}{RAFT-Stereo} & - & 1.17 & 1.37 & 0.483 & 0.507  \\
  & $t$=0.25 & \cellcolor{gray!20}1.11 & \cellcolor{gray!20}1.34 & \cellcolor{gray!20}0.478 & \cellcolor{gray!20}0.501 \\
  & $t$=0.5 & \cellcolor{gray!20}1.14 & 1.38 & \cellcolor{gray!20}0.479 & \cellcolor{gray!20}0.506 \\
  & $t$=0.75 & \cellcolor{gray!20}\textbf{1.07} & \cellcolor{gray!20}1.31 & \cellcolor{gray!20}\textbf{0.472} & \cellcolor{gray!20}\textbf{0.499} \\
  & $t$=1.0 & \cellcolor{gray!20}\textbf{1.07} & \cellcolor{gray!20}\textbf{1.29} & \cellcolor{gray!20}0.474 & \cellcolor{gray!20}\textbf{0.499} \\
  \hline
  \multirow{5}{*}{IGEV-Stereo} & - & 0.99 & 1.20 & 0.448 & 0.472 \\
  & $t$=0.25 & \cellcolor{gray!20}\textbf{0.96} & \cellcolor{gray!20}\textbf{1.16} & \cellcolor{gray!20}\textbf{0.445} & \cellcolor{gray!20}\textbf{0.468} \\
  & $t$=0.5 & \cellcolor{gray!20}\textbf{0.96} & \cellcolor{gray!20}1.17 & \cellcolor{gray!20}\textbf{0.445} & \cellcolor{gray!20}0.471 \\
  & $t$=0.75 & \cellcolor{gray!20}0.97 & \cellcolor{gray!20}\textbf{1.16} & \cellcolor{gray!20}0.446 & \cellcolor{gray!20}0.471 \\
  & $t$=1.0 & \cellcolor{gray!20}0.97 & \cellcolor{gray!20}1.17 & \cellcolor{gray!20}0.446 & \cellcolor{gray!20}0.469 \\
  \hline
  \multirow{5}{*}{SR-Stereo} & - & 0.98 & 1.20 & 0.443 & 0.471 \\
  & $t$=0.25 & \cellcolor{gray!20}\textbf{0.95} & \cellcolor{gray!20}\textbf{1.16} & \cellcolor{gray!20}\textbf{0.440} & \cellcolor{gray!20}0.468 \\
  & $t$=0.5 & \cellcolor{gray!20}0.97 & \cellcolor{gray!20}1.17 & \cellcolor{gray!20}0.442 & \cellcolor{gray!20}\textbf{0.465} \\
  & $t$=0.75 & \cellcolor{gray!20}0.97 & \cellcolor{gray!20}\textbf{1.16} & 0.443 & \cellcolor{gray!20}0.469 \\
  & $t$=1.0 & 0.98 & \cellcolor{gray!20}\textbf{1.16} & 0.446 & \cellcolor{gray!20}0.469 \\
  \hline
  \end{tabular}
\end{table}

\textit{Results on KITTI}:
Considering the limitation of the number of KITTI online leaderboard submissions, 
we have divided the KITTI training set into two parts, with a ratio of 4:1 
for fine-tuning the model and evaluation, respectively. 
The dataset provides sparse ground truth disparities obtained from lidar measurements. 
Notably, the upper regions of KITTI images primarily consist of sky and distant objects, 
where ground truth disparities are not available. Moreover, lidar performs poorly 
in the edge regions of objects, resulting in a lack of ground truth disparities 
for pixels in these areas. These factors lead to a mismatch between the 
accuracy ranking and the visualization on the KITTI online leaderboard. 
To ensure a more objective evaluation of the effectiveness of the proposed DAPE, 
we present the experimental results of DAPE on different models from both 
quantitative performance and visualization.

\begin{figure}[t]
  \centering
  \includegraphics[width=0.49\textwidth]{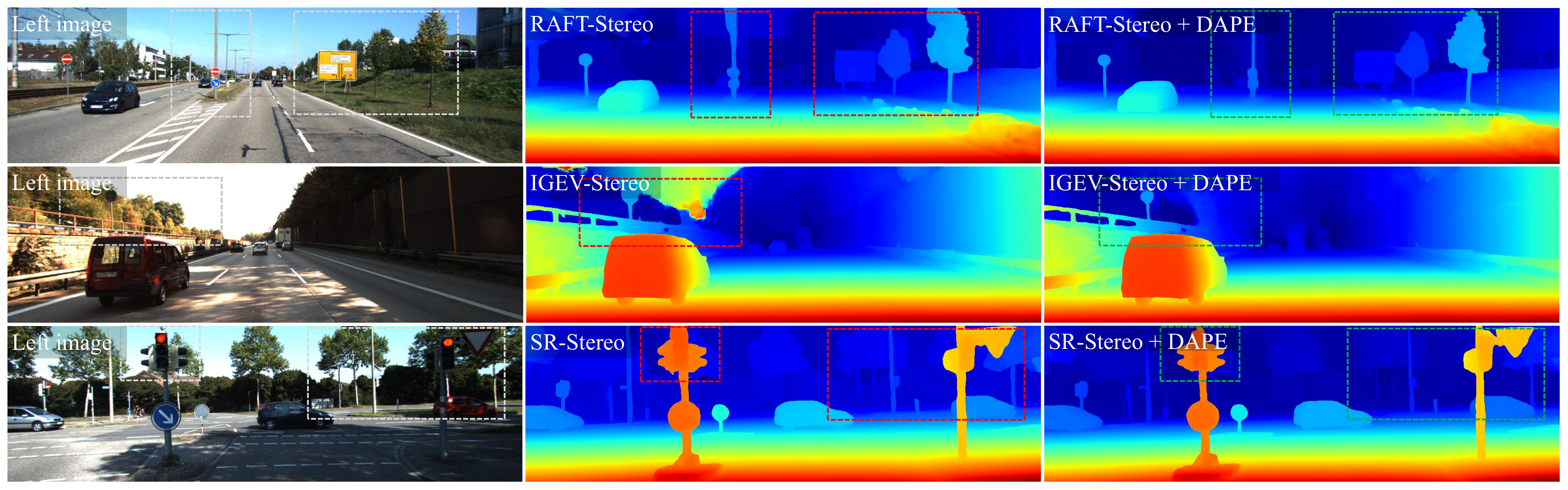}
  \caption{Qualitative disparity estimation results of DAPE on KITTI test set. 
  All methods run 15 disparity updates during inference. 
  For RAFT-Stereo, the threshold $t$ used for DAPE is 1, while 
  for the other two models, it is 0.25.}
  \label{fig:13}
\end{figure}

\begin{table}[t]
  \caption{Domain adaptation evaluation of DAPE on Middbury 2014. 
  All methods are fine-tuned on the sparsified ground 
  truth and evaluated on the original ground truth.
  Due to the large disparity range of Middlebury 2014, 
  RAFT-Stereo runs 32 disparity updates during inference, 
  while the other two models run 15 disparity updates.
  We use the 2-pixel error as the evaluation metric.
  Gray: performance is improved after using DAPE. \textbf{Bold}: Best.}
  \label{tab:dape_Middbury}
  \centering
  \begin{tabular}{|c|l|ccc|}
  \hline
  Methods & DAPE & Full & Half & Quarter \\
  \hline
  \multirow{5}{*}{RAFT-Stereo} & - & 12.85 & 8.14 & 7.56 \\
  & $t$=0.25 & \cellcolor{gray!20}11.71 & \cellcolor{gray!20}7.89 & \cellcolor{gray!20}7.27 \\
  & $t$=0.5 & \cellcolor{gray!20}11.56 & \cellcolor{gray!20}7.89 & \cellcolor{gray!20}7.02 \\
  & $t$=0.75 & \cellcolor{gray!20}11.58 & \cellcolor{gray!20}\textbf{7.81} &  \cellcolor{gray!20}7.00 \\
  & $t$=1.0 & \cellcolor{gray!20}\textbf{11.47} & \cellcolor{gray!20}7.84 & \cellcolor{gray!20}\textbf{6.87} \\
  \hline
  \multirow{5}{*}{IGEV-Stereo} & - & 11.72 & 5.14 & 5.00 \\
  & $t$=0.25 & \cellcolor{gray!20}11.69 & \cellcolor{gray!20}4.93 & \cellcolor{gray!20}4.99 \\
  & $t$=0.5 & \cellcolor{gray!20}\textbf{11.38} & \cellcolor{gray!20}4.93 & \cellcolor{gray!20}4.98 \\
  & $t$=0.75 & \cellcolor{gray!20}11.49 & \cellcolor{gray!20}\textbf{4.82} & \cellcolor{gray!20}4.86 \\
  & $t$=1.0 & \cellcolor{gray!20}11.53 & \cellcolor{gray!20}4.92 & \cellcolor{gray!20}\textbf{4.85} \\
  \hline
  \multirow{5}{*}{SR-Stereo} & - & 11.06 & 4.99 & 4.65 \\
  & $t$=0.25 & \cellcolor{gray!20}10.97 & \cellcolor{gray!20}\textbf{4.81} & \cellcolor{gray!20}4.62 \\
  & $t$=0.5 & \cellcolor{gray!20}\textbf{10.90} & 5.00 & \cellcolor{gray!20}4.61 \\
  & $t$=0.75 & \cellcolor{gray!20}11.02 & \cellcolor{gray!20}4.97 & \cellcolor{gray!20}4.58 \\
  & $t$=1.0 & \cellcolor{gray!20}11.01 & \cellcolor{gray!20}4.97 & \cellcolor{gray!20}\textbf{4.56} \\
  \hline
  \end{tabular}
\end{table}

As shown in Table~\ref{tab:dape_kitti}, the implementation of DAPE with most threshold 
values leads to performance improvements in both non-occluded and 
overall regions for all three models. Notably, the RAFT-Stereo model benefits 
significantly from higher-density edge pseudo-labels. Despite the presence 
of some erroneous labels in high-density edge pseudo-labels, they provide 
more comprehensive edge information, which effectively complements the edge 
disparity update guidance in RAFT-Stereo, especially considering its 
initial disparity is set to zero. Figure~\ref{fig:13} showcases the qualitative 
results of DAPE on different models. Alongside enhancing the model's 
performance in detailed regions, the proposed DAPE successfully mitigates 
disparity anomalies in textureless sky areas of the images.

\textit{Results on Middbury}: 
We use the additional 13 image pairs provided by Middlebury 2014 to 
fine-tune the models, and evaluate the domain adaptation performance 
using the original 15 training image pairs. Similar to the 
experiments conducted on ETH3D, we employ sparsified ground 
truth during the fine-tuning process, while relying on the 
original ground truth for evaluation, as shown in Figure~\ref{fig:11}.

To verify the performance of DAPE at different resolutions, 
we evaluate it on Middlebury 2014 with three different resolutions. 
The qualitative results of DAPE on Middlebury 2014 are presented in Table~\ref{tab:dape_Middbury}. 
Consistent with the conclusions from the previous two datasets, the 
implementation of DAPE with most threshold values leads to performance 
improvements of the three models. This outcome further confirms 
the generalizability and robustness of DAPE across different scenarios.

\section{Conclusion}
In this paper, we propose a novel stereo method, SR-Stereo, 
and a novel fine-tuning framework, DAPE. SR-Stereo overcomes 
the domain discrepancies by predicting multiple 
range-controlled disparity clips to achieve better generalization 
performance and domain adaptation performance. Furthermore, 
for domain adaptation on sparse ground truth, the proposed DAPE 
uses generated edge pseudo-labels to provide additional supervision 
to the fine-tuned model. We have conducted extensive experiments 
on Sceneflow, KITTI, Middlebury, and ETH3D, and effectively 
demonstrated the advancements of SR-Stereo and the effectiveness 
of DAPE. Compared to the baseline, the proposed SR-Stereo achieves 
both better cross-domain performance and in-domain performance.
% More importantly, although the original motivation of 
% the proposed SR-Stereo is to achieve better cross-domain performance, 
% it also effectively improves the in-domain performance of the baseline. 
This result is in complete contrast to many generalized stereo methods, 
which tend to have poor in-domain performance. In the future, we 
plan to implement a stereo method based entirely on stepwise 
regression by designing more components related to stepwise 
regression architecture.

% \section*{Acknowledgments}
% This should be a simple paragraph before the References to thank those individuals and institutions who have supported your work on this article.



% \end{IEEEbiography}

\vspace{11pt}
\begin{IEEEbiographynophoto}{Weiqing Xiao} received the B.S. degree from 
  SHENYUAN Honors College of Beihang University in 2022. 
  He is currently pursuing the master's degree in School of the School of 
  Electronic and Information Engineering, Beihang University. His research interests include 
  computer vision and 3D vision.
\end{IEEEbiographynophoto}

\begin{IEEEbiographynophoto}{Wei Zhao} received the B.S., M.S., and Ph.D.
  degrees from the School of Automatic Control, Northwestern Polytechnical University, Xi’an,
  China. She was a Post-Doctoral Researcher with Beihang University, where she is currently a Full
  Professor. Her main research interests are digital image processing, automatic target
  recognition, information fusion.
\end{IEEEbiographynophoto}

\vfill


\begin{thebibliography}{1}
\bibliographystyle{IEEEtran}

\bibitem{zhang2013real}
Y.~Zhang, Z.~Xiong, Z.~Yang, and F.~Wu, ``Real-time scalable depth sensing with hybrid structured light illumination,'' \emph{IEEE Transactions on Image Processing}, vol.~23, no.~1, pp. 97--109, 2013.

% tits
\bibitem{liu2024guard}
Y.~Liu, X.~Zhang, Y.~Luo, Q.~Hao, J.~Su, and G.~Cai, ``Guard-net: Lightweight stereo matching network via global and uncertainty-aware refinement for autonomous driving,'' \emph{IEEE Transactions on Intelligent Transportation Systems}, vol.~25, no.~8, pp. 10\,260--10\,273, 2024.

% tits
\bibitem{feng2018lane}
Z.~Feng, M.~Li, M.~Stolz, M.~Kunert, and W.~Wiesbeck, ``Lane detection with a high-resolution automotive radar by introducing a new type of road marking,'' \emph{IEEE Transactions on Intelligent Transportation Systems}, vol.~20, no.~7, pp. 2430--2447, 2018.

\bibitem{li2021revisiting}
Li, Z., Liu, X., Drenkow, N., Ding, A., Creighton, F.X., Taylor, R.H., Unberath, M.: Revisiting stereo depth estimation from a sequence-to-sequence perspective with transformers. In: Proceedings of the IEEE/CVF international conference on computer vision. pp. 6197--6206 (2021)

\bibitem{chang2018pyramid}
Chang, J.R., Chen, Y.S.: Pyramid stereo matching network. In: Proceedings of the IEEE conference on computer vision and pattern recognition. pp. 5410--5418 (2018)

\bibitem{guo2019group}
Guo, X., Yang, K., Yang, W., Wang, X., Li, H.: Group-wise correlation stereo network. In: Proceedings of the IEEE/CVF conference on computer vision and pattern recognition. pp. 3273--3282 (2019)

\bibitem{xu2022attention}
Xu, G., Cheng, J., Guo, P., Yang, X.: Attention concatenation volume for accurate and efficient stereo matching. In: Proceedings of the IEEE/CVF Conference on Computer Vision and Pattern Recognition. pp. 12981--12990 (2022)

\bibitem{li2022practical}
Li, J., Wang, P., Xiong, P., Cai, T., Yan, Z., Yang, L., Liu, J., Fan, H., Liu, S.: Practical stereo matching via cascaded recurrent network with adaptive correlation. In: Proceedings of the IEEE/CVF conference on computer vision and pattern recognition. pp. 16263--16272 (2022)

\bibitem{lipson2021raft}
Lipson, L., Teed, Z., Deng, J.: Raft-stereo: Multilevel recurrent field transforms for stereo matching. In: 2021 International Conference on 3D Vision (3DV). pp. 218--227. IEEE (2021)

\bibitem{xu2023iterative}
Xu, G., Wang, X., Ding, X., Yang, X.: Iterative geometry encoding volume for stereo matching. In: Proceedings of the IEEE/CVF Conference on Computer Vision and Pattern Recognition. pp. 21919--21928 (2023)

\bibitem{wang2024selective}
X.~Wang, G.~Xu, H.~Jia, and X.~Yang, ``Selective-stereo: Adaptive frequency information selection for stereo matching,'' \emph{arXiv preprint arXiv:2403.00486}, 2024.

\bibitem{weinzaepfel2023croco}
Weinzaepfel, P., Lucas, T., Leroy, V., Cabon, Y., Arora, V., Br{\'e}gier, R., Csurka, G., Antsfeld, L., Chidlovskii, B., Revaud, J.: Croco v2: Improved cross-view completion pre-training for stereo matching and optical flow. In: Proceedings of the IEEE/CVF International Conference on Computer Vision. pp. 17969--17980 (2023)

\bibitem{zhao2023high}
Zhao, H., Zhou, H., Zhang, Y., Chen, J., Yang, Y., Zhao, Y.: High-frequency stereo matching network. In: Proceedings of the IEEE/CVF Conference on Computer Vision and Pattern Recognition. pp. 1327--1336 (2023)

% \bibitem{shen2021cfnet}
% Shen, Z., Dai, Y., Rao, Z.: Cfnet: Cascade and fused cost volume for robust stereo matching. In: Proceedings of the IEEE/CVF Conference on Computer Vision and Pattern Recognition. pp. 13906--13915 (2021)

\bibitem{shen2023digging}
Shen, Z., Song, X., Dai, Y., Zhou, D., Rao, Z., Zhang, L.: Digging into uncertainty-based pseudo-label for robust stereo matching. IEEE Transactions on Pattern Analysis and Machine Intelligence  (2023)

\bibitem{zhang2020domain}
Zhang, F., Qi, X., Yang, R., Prisacariu, V., Wah, B., Torr, P.: Domain-invariant stereo matching networks. In: Computer Vision--ECCV 2020: 16th European Conference, Glasgow, UK, August 23--28, 2020, Proceedings, Part II 16. pp. 420--439. Springer (2020)

\bibitem{mayer2016large}
Mayer, N., Ilg, E., Hausser, P., Fischer, P., Cremers, D., Dosovitskiy, A., Brox, T.: A large dataset to train convolutional networks for disparity, optical flow, and scene flow estimation. In: Proceedings of the IEEE conference on computer vision and pattern recognition. pp. 4040--4048 (2016)

\bibitem{scharstein2014high}
Scharstein, D., Hirschm{\"u}ller, H., Kitajima, Y., Krathwohl, G., Ne{\v{s}}i{\'c}, N., Wang, X., Westling, P.: High-resolution stereo datasets with subpixel-accurate ground truth. In: Pattern Recognition: 36th German Conference, GCPR 2014, M{\"u}nster, Germany, September 2-5, 2014, Proceedings 36. pp. 31--42. Springer (2014)

\bibitem{menze2015object}
Menze, M., Geiger, A.: Object scene flow for autonomous vehicles. In: Proceedings of the IEEE conference on computer vision and pattern recognition. pp. 3061--3070 (2015)

\bibitem{geiger2012we}
Geiger, A., Lenz, P., Urtasun, R.: Are we ready for autonomous driving? the kitti vision benchmark suite. In: 2012 IEEE conference on computer vision and pattern recognition. pp. 3354--3361. IEEE (2012)

% tits
\bibitem{dong2023egfnet}
S.~Dong, W.~Zhou, C.~Xu, and W.~Yan, ``Egfnet: Edge-aware guidance fusion network for rgb--thermal urban scene parsing,'' \emph{IEEE Transactions on Intelligent Transportation Systems}, 2023.

% tits
\bibitem{cao2022pavement}
T.~Cao, Y.~Wang, and S.~Liu, ``Pavement crack detection based on 3d edge representation and data communication with digital twins,'' \emph{IEEE Transactions on Intelligent Transportation Systems}, vol.~24, no.~7, pp. 7697--7706, 2022.

% tits
\bibitem{guo2021barnet}
J.-M. Guo, H.~Markoni, and J.-D. Lee, ``Barnet: Boundary aware refinement network for crack detection,'' \emph{IEEE Transactions on Intelligent Transportation Systems}, vol.~23, no.~7, pp. 7343--7358, 2021.

\bibitem{zhang2019ga}
Zhang, F., Prisacariu, V., Yang, R., Torr, P.H.: Ga-net: Guided aggregation net for end-to-end stereo matching. In: Proceedings of the IEEE/CVF Conference on Computer Vision and Pattern Recognition. pp. 185--194 (2019)

\bibitem{zhang2020adaptive}
Zhang, Y., Chen, Y., Bai, X., Yu, S., Yu, K., Li, Z., Yang, K.: Adaptive unimodal cost volume filtering for deep stereo matching. In: Proceedings of the AAAI Conference on Artificial Intelligence. vol.~34, pp. 12926--12934 (2020)

% tits
\bibitem{zeng2021deep}
K.~Zeng, Y.~Wang, Q.~Zhu, J.~Mao, and H.~Zhang, ``Deep progressive fusion stereo network,'' \emph{IEEE Transactions on Intelligent Transportation Systems}, vol.~23, no.~12, pp. 25\,437--25\,447, 2021.


\bibitem{teed2020raft}
Teed, Z., Deng, J.: Raft: Recurrent all-pairs field transforms for optical flow. In: Computer Vision--ECCV 2020: 16th European Conference, Glasgow, UK, August 23--28, 2020, Proceedings, Part II 16. pp. 402--419. Springer (2020)

\bibitem{cho2014learning}
Cho, K., Van~Merri{\"e}nboer, B., Gulcehre, C., Bahdanau, D., Bougares, F., Schwenk, H., Bengio, Y.: Learning phrase representations using rnn encoder-decoder for statistical machine translation. arXiv preprint arXiv:1406.1078  (2014)

\bibitem{graves2012long}
Graves, A., Graves, A.: Long short-term memory. Supervised sequence labelling with recurrent neural networks pp. 37--45 (2012)

\bibitem{sandler2018mobilenetv2}
M.~Sandler, A.~Howard, M.~Zhu, A.~Zhmoginov, and L.-C. Chen, ``Mobilenetv2: Inverted residuals and linear bottlenecks,'' in \emph{Proceedings of the IEEE conference on computer vision and pattern recognition}, 2018, pp. 4510--4520.

\bibitem{krizhevsky2017imagenet}
A.~Krizhevsky, I.~Sutskever, and G.~E. Hinton, ``Imagenet classification with deep convolutional neural networks,'' \emph{Communications of the ACM}, vol.~60, no.~6, pp. 84--90, 2017.

\bibitem{he2016deep}
He, K., Zhang, X., Ren, S., Sun, J.: Deep residual learning for image recognition. In: Proceedings of the IEEE conference on computer vision and pattern recognition. pp. 770--778 (2016)

\bibitem{schops2017multi}
Schops, T., Schonberger, J.L., Galliani, S., Sattler, T., Schindler, K., Pollefeys, M., Geiger, A.: A multi-view stereo benchmark with high-resolution images and multi-camera videos. In: Proceedings of the IEEE conference on computer vision and pattern recognition. pp. 3260--3269 (2017)

\bibitem{chen2024mocha}
Z.~Chen, W.~Long, H.~Yao, Y.~Zhang, B.~Wang, Y.~Qin, and J.~Wu, ``Mocha-stereo: Motif channel attention network for stereo matching,'' in \emph{Proceedings of the IEEE/CVF Conference on Computer Vision and Pattern Recognition}, 2024, pp. 27\,768--27\,777.

\bibitem{guan2024neural}
T.~Guan, C.~Wang, and Y.-H. Liu, ``Neural markov random field for stereo matching,'' in \emph{Proceedings of the IEEE/CVF Conference on Computer Vision and Pattern Recognition}, 2024, pp. 5459--5469.

\bibitem{xu2024adaptive}
P.~Xu, Z.~Xiang, C.~Qiao, J.~Fu, and T.~Pu, ``Adaptive multi-modal cross-entropy loss for stereo matching,'' in \emph{Proceedings of the IEEE/CVF Conference on Computer Vision and Pattern Recognition}, 2024, pp. 5135--5144.

\bibitem{zhang2024robust}
J.~Zhang, J.~Li, L.~Huang, X.~Yu, L.~Gu, J.~Zheng, and X.~Bai, ``Robust synthetic-to-real transfer for stereo matching,'' in \emph{Proceedings of the IEEE/CVF Conference on Computer Vision and Pattern Recognition}, 2024, pp. 20\,247--20\,257.

\bibitem{liu2022graftnet}
B.~Liu, H.~Yu, and G.~Qi, ``Graftnet: Towards domain generalized stereo matching with a broad-spectrum and task-oriented feature,'' in \emph{Proceedings of the IEEE/CVF conference on computer vision and pattern recognition}, 2022, pp. 13\,012--13\,021.

\bibitem{zhang2022revisiting}
J.~Zhang, X.~Wang, X.~Bai, C.~Wang, L.~Huang, Y.~Chen, L.~Gu, J.~Zhou, T.~Harada, and E.~R. Hancock, ``Revisiting domain generalized stereo matching networks from a feature consistency perspective,'' in \emph{Proceedings of the IEEE/CVF Conference on Computer Vision and Pattern Recognition}, 2022, pp. 13\,001--13\,011.

\bibitem{chang2023domain}
T.~Chang, X.~Yang, T.~Zhang, and M.~Wang, ``Domain generalized stereo matching via hierarchical visual transformation,'' in \emph{Proceedings of the IEEE/CVF Conference on Computer Vision and Pattern Recognition}, 2023, pp. 9559--9568.

\bibitem{rao2023masked}
Z.~Rao, B.~Xiong, M.~He, Y.~Dai, R.~He, Z.~Shen, and X.~Li, ``Masked representation learning for domain generalized stereo matching,'' in \emph{Proceedings of the IEEE/CVF Conference on Computer Vision and Pattern Recognition}, 2023, pp. 5435--5444.

\bibitem{song2020edgestereo}
X.~Song, X.~Zhao, L.~Fang, H.~Hu, and Y.~Yu, ``Edgestereo: An effective multi-task learning network for stereo matching and edge detection,'' \emph{International Journal of Computer Vision}, vol. 128, no.~4, pp. 910--930, 2020.

% ----------------no use
\end{thebibliography}
\end{document}